\documentclass[lettersize,journal]{IEEEtran}
\ifCLASSOPTIONcompsoc
  \usepackage[nocompress]{cite}
\else
  \usepackage{cite}
\fi
\ifCLASSINFOpdf
\else
\fi
\usepackage{amsmath,amsfonts}
\usepackage{algorithmic}
\usepackage{array}
\usepackage{textcomp}
\usepackage{stfloats}
\usepackage{url}
\usepackage{verbatim}
\usepackage{graphicx}
\usepackage{amssymb}
\usepackage{bm}
\usepackage{multirow}
\usepackage{color}
\usepackage{cite}
\usepackage{tabularx}
\usepackage{float}
\usepackage{fdsymbol}
\usepackage[switch]{lineno}
\usepackage{hyperref}
\usepackage{bbding}
\usepackage{booktabs}
\usepackage{subcaption}

\begin{document}

\title{Point Cloud Compression and Objective Quality Assessment: A Survey}

\markboth{Journal of \LaTeX\ Class Files,~Vol.~14, No.~8, August~2015}%
{Shell \MakeLowercase{\textit{et al.}}: Bare Demo of IEEEtran.cls for Computer Society Journals}

\author{Yiling Xu,~\IEEEmembership{Member,~IEEE}, Yujie Zhang, Shuting Xia, Kaifa Yang, He Huang, Ziyu Shan, Wenjie Huang, \\Qi Yang,  ~\IEEEmembership{Member,~IEEE}, Le Yang
\thanks{This paper is supported in part by National Natural Science Foundation of China (62371290), National Key R\&D Program of China (2024YFB2907204), the Fundamental Research Funds for the Central Universities of China, and STCSM under Grant (22DZ2229005). The corresponding author is Yiling Xu(e-mail: yl.xu@sjtu.edu.cn).  }
\thanks{Y. Xu, Y. Zhang, S. Xia, K. Yang, H. Huang, Z. Shan, W. Huang are with the Cooperative Medianet Innovation Center, Shanghai Jiao Tong University, Shanghai, China.}
\thanks{Q. Yang is with University of Missouri–Kansas City, Kansa city, America. }
\thanks{L. Yang is with the Department of Electrical and Computer Engineering, University of Canterbury, Christchurch, New Zealand. }

}



\IEEEtitleabstractindextext{%
\begin{abstract}
The rapid growth of 3D point cloud data, driven by applications in autonomous driving, robotics, and immersive environments, has led to criticals demand for efficient compression and quality assessment techniques. Unlike traditional 2D media, point clouds present unique challenges due to their irregular structure, high data volume, and complex attributes. This paper provides a comprehensive survey of recent advances in point cloud compression (PCC) and point cloud quality assessment (PCQA), emphasizing their significance for real-time and perceptually relevant applications. We analyze a wide range of handcrafted and learning-based PCC algorithms, along with objective PCQA metrics. By benchmarking representative methods on emerging datasets, we offer detailed comparisons and practical insights into their strengths and limitations. Despite notable progress, challenges such as enhancing visual fidelity, reducing latency, and supporting multimodal data remain. This survey outlines future directions, including hybrid compression frameworks and advanced feature extraction strategies, to enable more efficient, immersive, and intelligent 3D applications.

\end{abstract}

\begin{IEEEkeywords}
Point cloud, data compression, quality assessment, large-scale performance evaluation 
\end{IEEEkeywords}}

\maketitle

\IEEEdisplaynontitleabstractindextext

% \fi
\IEEEpeerreviewmaketitle

\section{Introduction}
A point cloud (PC) refers to a collection of points in a 3D coordinate system, typically composed of their spatial coordinates $(x,y,z)$ and optionally associated with additional attributes such as color, intensity, or normal vectors. They are widely used to represent the surface and geometric structure of real-world objects and scenes. PCs can be acquired through devices such as depth sensors, RGB cameras, and LiDAR or synthesized via computational methods such as 3D reconstruction from multi-view images and sampling from polygonal meshes. This versatility gives PCs the essential role in various applications, including autonomous driving, augmented/virtual reality (AR/VR), robotics, and digital twins. Meanwhile, the demand for efficient processing, storage, and transmission of PCs has grown significantly.

The massive volume of PCs is a major obstacle that restricts their application. For instance, in a dynamic human point cloud sequence, a single frame can contain up to 1 million points, with each point’s geometry bit depth of 10 and color attribute bit depth of 8 per channel. Considering a frame rate of 30 frames per second (FPS), uncompressed 10-second dynamic point cloud data can reach approximately 1.9 gigabytes (GB). Thus, point cloud compression (PCC) is indispensable for efficient streaming and rendering, ensuring robust downstream processing and immersive user experiences. Unlike conventional 2D media with a regular grid structure, 3D PCs are characterized by e.g., sparse and non-uniform distribution, incompleteness, and noise susceptibility. These necessitate the design of specialized compression algorithms to effectively exploit spatial-temporal properties of PCs. 

% Meanwhile, real-time applications such as autonomous driving and AR/VR require low-latency transmission of high-quality point clouds. Efficient compression is indispensable for supporting efficient streaming and rendering, ensuring robust downstream processing and immersive user experiences.

After compression and other practical processing stages such as transmission and rendering, PCs will be inevitably subject to a variety of distortions, which usually impair the perceptual quality of the human visual system (HVS). To guide point cloud applications targeted at human vision, point cloud quality assessment (PCQA)
has been widely explored. It can be further categorized into subjective PCQA and objective PCQA. Although subjective PCQA can provide accurate predictions, its application is limited due to the high cost in terms of time and cost. On the contrary, objective PCQA intends to design automatic approaches called PCQA metrics that can predict point cloud quality consistent with subjective evaluation results. This is more convenient in practice due to lower cost and faster speed. PCQA metrics play a vital role in quality of experience (QoE)-oriented data processing. For instance, for PCC or transmission, PCQA metrics can be used to derive constraints to achieve a better trade-off between quality and compression rate or transmission bandwidth. Moreover, PCQA metrics can be included as loss functions to guide model optimization in learning-based PC generation tasks, such as spatial upsampling and completion. Therefore, it is highly desired to develop reliable and effective objective PCQA metrics that are  well aligned with human perception.

Toward facilitating a better understanding the current progress
on PCC and PCQA, we attempt to conduct
a comprehensive overview of the past and recent advances. The key contributions include:

\begin{itemize}
    \item We perform a comprehensive overview of both PCC and PCQA, providing a detailed taxonomy as well as a nuanced analysis of their evolution.
    \item We conduct experiments to compare representative PCC/PCQA models on multiple public databases, and we provide effective insights into the future development.
    \item We discuss practical applications, some of which have been deployed at very large commercial scales, remaining core challenges,
    and future opportunities for improving and applying PCC and PCQA models.
\end{itemize}

The structure of this paper is as follows. Section \ref{sec:PCC}
reviews the existing
work of PCC and compares
the performance of representative methods on emerging
databases. Section \ref{sec:PCQA} provides a survey of existing
metrics for PCQA and conducts experiments on multiple publicly available databases. Section \ref{sec:application} discusses practical applications
and challenges, and Section \ref{sec:conclusion} concludes with
forward-looking insights.

\section{Point Cloud Compression}\label{sec:PCC}
% {\bf Outline: }The section introduces:
% \begin{itemize}
%     \item Handcrafted Point Cloud Compression:
%     \begin{itemize}
%         \item Video-based Point Cloud Compression
%         \item Geometry-based Point Cloud Compression
%     \end{itemize}
%     \item Learning-based Point Cloud Compression:
%     \begin{itemize}
%         \item Point-based Point Cloud Compression
%         \item Voxel-based Point Cloud Compression
%         \item Tree-based Point Cloud Compression
%         \item Heterogeneity-based Point Cloud Compression 
%     \end{itemize}
%     \item Performance evaluation of existing methods under different conditions
% \end{itemize}

\subsection{Handcrafted Point Cloud Compression}
\begin{figure}[t]
    \centering
    \includegraphics[width=0.9\linewidth]{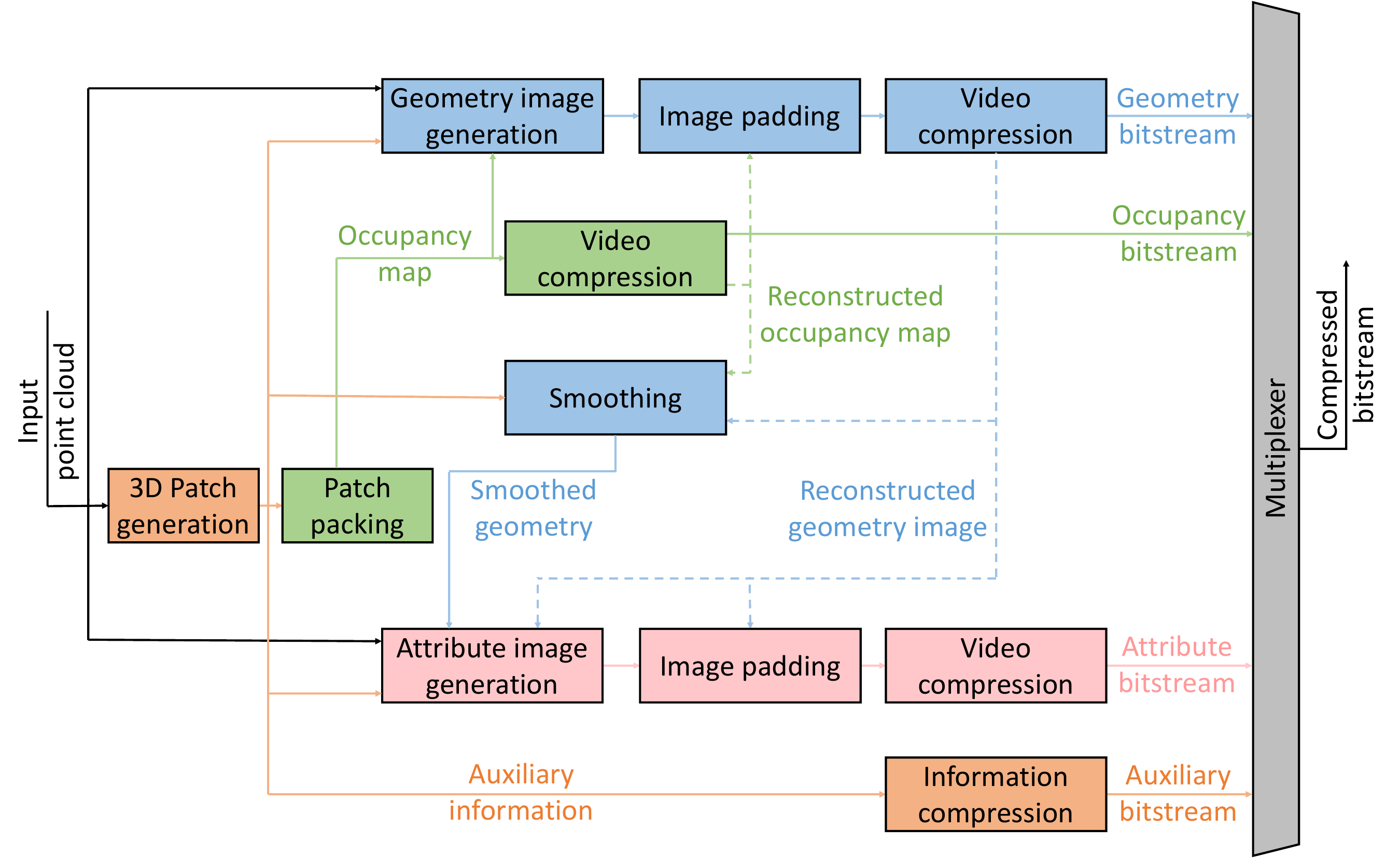}
    \caption{The encoding framework of V-PCC.}
    \label{fig:encoding framework of vpcc}
    \vspace{-0.6mm}
\end{figure}
\begin{figure}[t]
    \centering
    \includegraphics[width=0.9\linewidth]{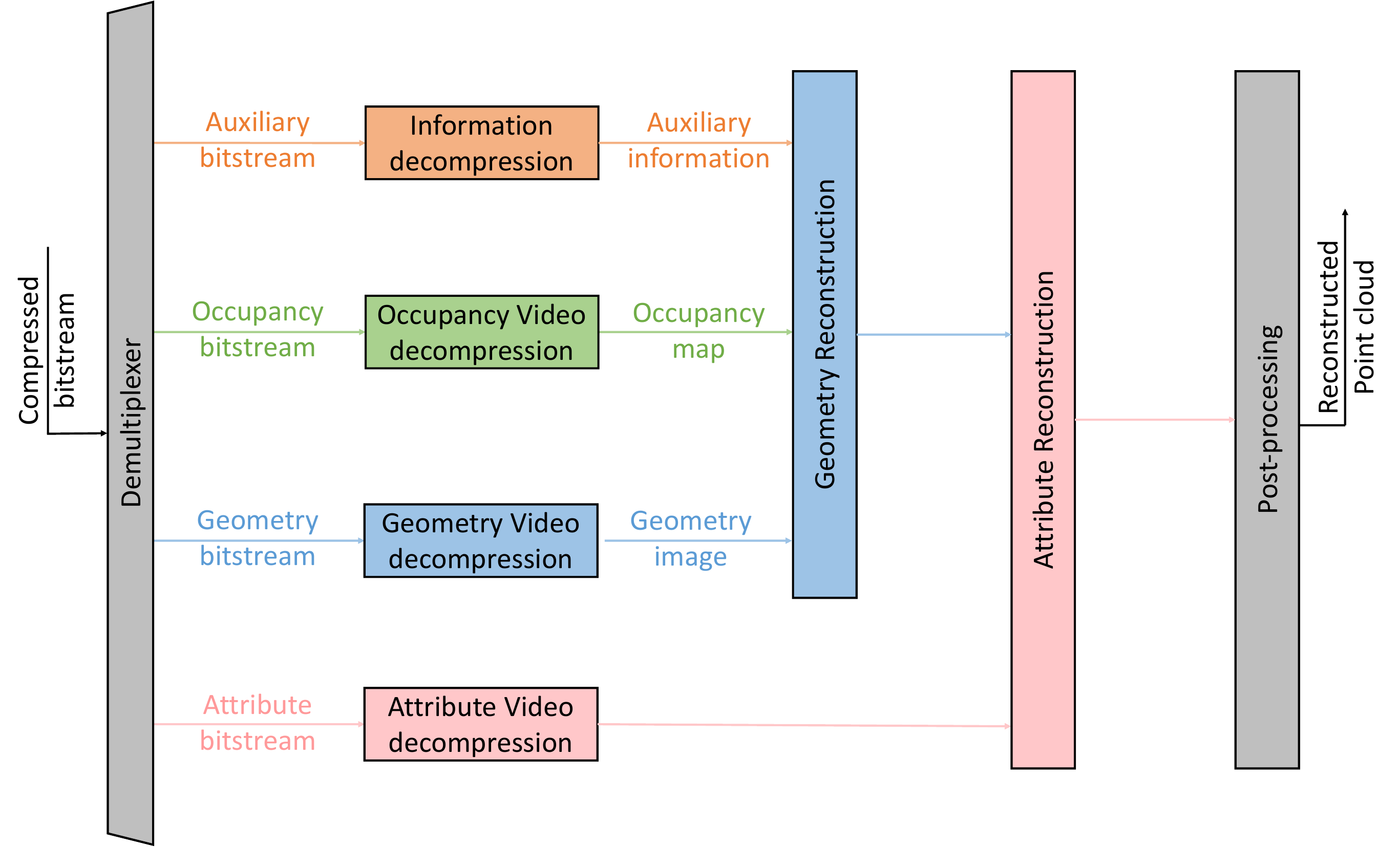}
    \caption{The decoding framework of V-PCC.}
    \label{fig:decoding framework of vpcc}
\end{figure}

\subsubsection{2D Projection-based Point Cloud Compression} 
2D projection-based point cloud compression methods first segment 3D point clouds into patches and then project them onto 2D planes for efficient representation. Video-based point cloud compression (V-PCC) developed by MPEG is a prominent approach, which uses established video codecs such as HEVC \cite{pcc_HEVC}. As illustrated in Fig. \ref{fig:encoding framework of vpcc}, the encoding process of V-PCC includes three key components: the occupancy map, geometry map, and attribute map, which preserve crucial spatial and attribute information. 
Specifically, the occupancy map is a binary representation that identifies regions in the 2D projection space. The geometry map provides spatial positions by mapping the original 3D point cloud onto a 2D grid based on the occupancy map, with depth values assigned to occupied regions. To reduce artifacts and enhance compression efficiency, padding is applied to unoccupied areas, creating smooth transitions at patch boundaries. This padding process uses interpolation techniques or synthesized data to mitigate discontinuities between adjacent patches.
Similarly, the attribute map encodes additional properties of the point cloud, such as color, reflectance, or other per-point attributes. Padding and smoothing techniques are also applied, where attribute values are extrapolated into unoccupied regions using interpolation or diffusion-based techniques.
The three maps are compressed separately using a 2D video codec, while auxiliary information such as patch position, size, and orientation is losslessly compressed into a separate bitstream for accurate 3D reconstruction. V-PCC decoding process is the inverse process of encoding, as shown in Fig.~\ref{fig:decoding framework of vpcc}.
By efficiently mapping 3D point clouds into a structured 2D representation and utilizing advanced video compression techniques, V-PCC supports both lossy and lossless compression modes.

Although V-PCC has achieved significant improvements in compression efficiency, it still has difficulties in maintaining data fidelity, handling redundancy, and compressing dynamic 3D data. To address these challenges, a variety of optimization techniques have been proposed to enhance V-PCC’s performance across various aspects of the encoding process. 
For example, to reduce data loss during projection, Li \textit{et al.} \cite{pcc_projection1} introduced an efficient projected frame padding. Eurico \textit{et al.} \cite{pcc_projection2} proposed an adaptive plane projection and Yu \textit{et al.} \cite{pcc_projection3} presented a regularized projection-based geometry compression.

In addition to utilizing video codecs, recent advancements in dynamic V-PCC have brought several innovative approaches to improve both inter-frame compression efficiency and visual quality. Motion prediction techniques have been significantly enhanced through methods like 3D motion prediction and view-dependent compression \cite{pcc_dynamic1, pcc_dynamic2, pcc_dynamic5}. For motion-specific compression, keyframe-based geometry \cite{pcc_dynamic3, pcc_dynamic4, pcc_dynamic6} are proposed to compress time-varying 3D data. 

Quantization errors introduced during the encoding process can lead to compression artifacts. To mitigate this, sparse convolutional networks \cite{pcc_enhance1, pcc_enhance2} have been widely employed to learn embeddings, effectively modeling and removing quantization noise from the input point cloud. Furthermore, occupancy map-based rate distortion optimization techniques \cite{pcc_enhance3, pcc_enhance4} aim to improve the compression efficiency of well-characterized pixels within the occupancy map, leading to better overall quality.

In summary, V-PCC excels in compressing dynamic and dense point clouds, achieving high compression ratios and maintaining visual quality. However, its reliance on projection-based methods limits its performance on sparse point clouds and its ability to attain lossless compression. Future research should address these challenges to expand its applicability.

\subsubsection{3D Correlations-based Point Cloud Compression}
Another mainstream approach directly exploits the inherent correlation among 3D points. Within this framework, tree-based representations including octree and kd-tree play an important role in geometry compression. On the other hand, attribute compression contains two major categories: predictive residual coding and frequency-domain transformation.

\begin{figure}
    \centering
    \includegraphics[width=0.9\linewidth]{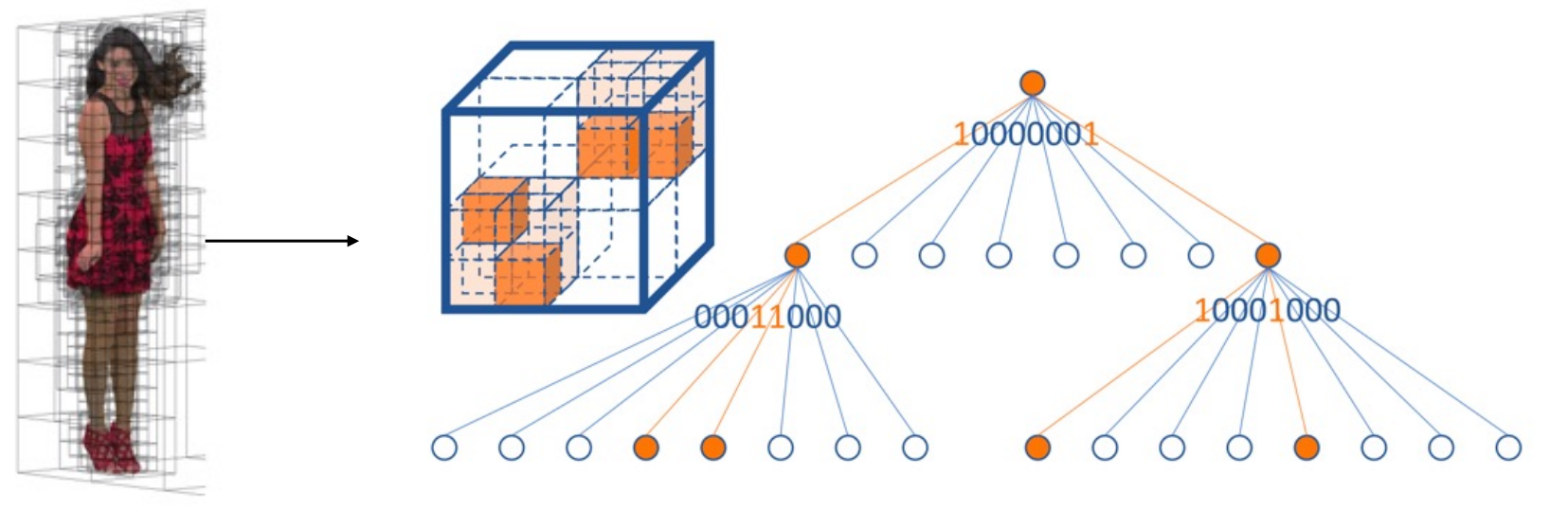}
    \caption{Illustration of octree-based point cloud geometry compression.}
    \label{fig:octree}
\end{figure}

\textbf{Tree-based Geometry Compression.}  
Octree\cite{pcc_octree} is a popular structure for point cloud geometry compression, as shown in Fig.~\ref{fig:octree}. It voxelizes the point cloud by quantizing point coordinates and recursively partitioning the occupied 3D space into 8 child nodes until reaching leaf nodes, and encodes each node's 8-bit child occupancy into the bitstream. The decoder can then reconstruct the point cloud with multi-level occupancy in a coarse-to-fine style. To expand octree-based coding, quad-tree and binary-tree \cite{pcc_qtbt1, pcc_qtbt2} serve as alternative for flexible partitions. Planar mode\cite{pcc_planar_mode} is designed for coplanar non-empty child nodes in a node volume and direct mode\cite{pcc_dct_mode} to directly code isolated points without recursive partition. Trisoup\cite{pcc_trisoup_1, pcc_trisoup_3} approximates a surface after a certain level of octree partition and is particularly suitable for solid point clouds.
As an efficient representation, octree is also widely adopted in the following learning-based point cloud compression frameworks, which we will give more details in Section \ref{sec:lr-pcc}.
% For a point with coordinate $(x,y,z)$, it is assigned to a voxel $(\lfloor \frac{x}{s}\rfloor, \lfloor \frac{y}{s}\rfloor, \lfloor \frac{z}{s}\rfloor)$, where $\lfloor \cdot \rfloor$ is the floor function, $s$ determines the representation precision.
% A voxel is a 3D equivalent of a pixel. It represents a small cubical unit in a 3D space. A voxel grid is a 3D array where each cell can be occupied or empty, depending on whether a point from the point cloud falls within its boundaries.
%One drawback of octree is that it requires voxelization, which inevitably introduces losses to the raw points.

\begin{figure}
    \centering
    \includegraphics[width=0.9\linewidth]{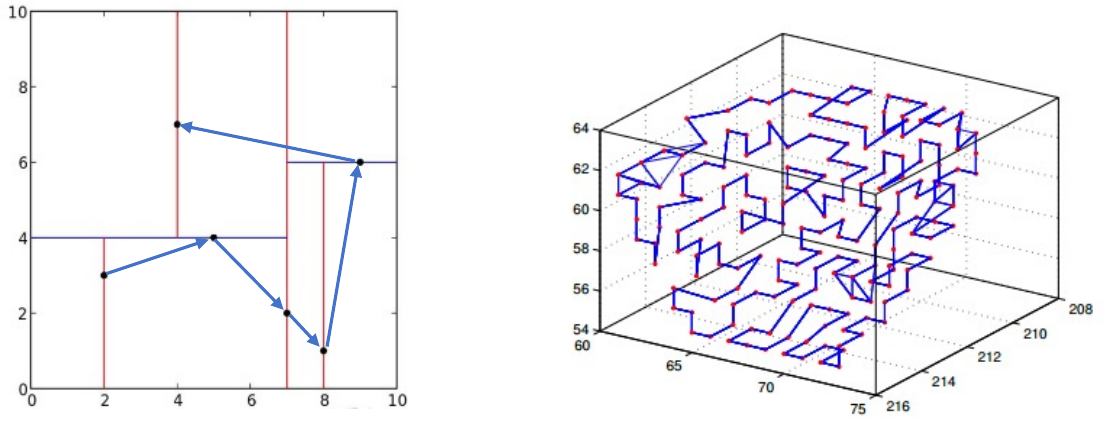}
    \caption{2D and 3D traversal order built by kd-tree.}
    \label{fig:predictive_tree}
    \vspace{-0.5cm}
\end{figure}

Kd-tree is another widely used tree structure for point cloud geometry compression, which partitions 3D space recursively based on the median values of coordinates. Kd-tree is efficient to find the nearest neighbors for a random point. Therefore, the traversal order can be built with the minimum distance residual criterion, as depicted in Fig.~\ref{fig:predictive_tree}. To expand kd-tree-based coding,
Gumhold \textit{et al.}\cite{pcc_predictive_tree} utilize kd-tree to construct a predictive tree, while Zhu \textit{et al.} \cite{pcc_Zhu2017LosslessPC} optimize the points' order by a Traveling Salesman Problem solver. Flynn \textit{et al.} \cite{pcc_predictive_tree_gpcc, pcc_predictive_tree_gpcc_angular} further enhance the prediction tree by introducing more predictors and an angular prior. However, the prediction tree is less efficient for dense point clouds due to the limited number of neighbors.

\begin{figure}[t]
    \centering
    \includegraphics[width=0.9\linewidth]{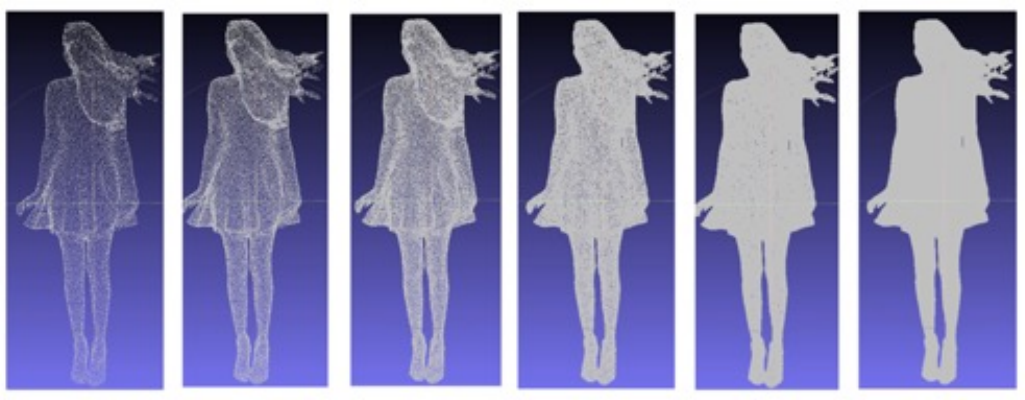}
    \caption{Example of LoDs. From left to right is $LoD(0)$ to $LoD(N)$.}
    \label{fig:lod}
    \vspace{-0.5cm}
\end{figure}

\textbf{Prediction-based Attribute Compression.} Prediction-based attribute compression relies on a traversal order similar to the kd-tree-based geometry compression, where the level of detail (LoD) order \cite{pcc_lod} is a typical example. Specifically, LoD divides a point cloud into hierarchical levels $R(0), R(1), \ldots, R(N)$ based on a series of descending point-to-point distance thresholds $\{d_{0}, d_{1}, \ldots, d_{N}\}$. The level of detail $j$, denoted as $LoD(j)$, is obtained by taking the union of all the refinement levels $R(0), R(1), …, R(j)$:
\begin{equation}
    LoD(j) = LoD(j-1)\cup R(j), j>0.
\end{equation}
Based on LoD, Mammou \textit{et al.} \cite{pcc_lifting} assign adaptive predictive weights for different levels. Wei\cite{pcc_morton_predict} and Chen \cite{pcc_hilbert_predict} propose Morton and Hilbert codes for traversal order. However, these methods share a common problem of the predictive tree-based geometry compression: coding efficiency is difficult to be further improved by utilizing more neighboring points.

\textbf{Transform-based Attribute Compression.} Inspired by the success of discrete cosine transformation (DCT) in 2D coding, point cloud attribute compression also applies frequency-domain transformations. Region adaptive hierarchical transform (RAHT) \cite{pcc_raht, pcc_sp_raht, pcc_integer_raht, pcc_dyadic_raht, pcc_Zhang2019A3H, pcc_skip_coding_raht} is a representative method, which depends on a hierarchical voxelized representation. At each resolution level, the transform is performed sequentially in $xyz$ directions. Each pair of occupied local sub-blocks is grouped to generate a DC coefficient and an AC coefficient.
%If an occupied sub-block has no pair neighbor in the grouping direction, no transform will be needed.
AC coefficients are quantized and entropy coded, while the DC coefficient is forwarded to the subsequent resolution level until the root. 

Alternatively, graph transform methods \cite{pcc_gft_1, pcc_gft_2, pcc_gft_3, pcc_gt_sdu, pcc_gt_pku_mm2024} process discrete points without voxelization. These methods first partition the point cloud into blocks, and then construct graphs with Euclidean distance based edge weights, followed by analyzing the graph Laplacian matrix to generate low- and high-frequency coefficients for optimal compression. Although graph transform achieves impressive compression efficiency, it results in high complexity because of graph construction and matrix operations. 

% \textbf{Standardization of Geometry-based Point Cloud Compression.} Based on the idea of directly exploiting 3D space correlation, MPEG is developing the Geometry-based Point Cloud Compression (G-PCC) standard framework. G-PCC offers multiple coding tools, including octree, predictive tree, and trisoup for geometry coding, LoD-based prediction and RAHT for attribute coding to adapt to various application scenarios.
% By October 2020, the first edition of G-PCC had reached the Final Draft International Standard (FDIS) stage\cite{pcc_gpcc_fdis}. At the virtual meeting in April 2022, it was decided to begin the second edition of the G-PCC standard, which mainly focuses on the removal of temporal redundancy.

% \begin{figure*}
%     \centering
%     \includegraphics[width=1\linewidth]{figures/aipcc_overview.pdf}
%     \caption{Chronological overview of the most relevant deep learning-based point cloud compression methods.}
%     \label{fig:timeline}
% \end{figure*}

\subsection{Learning-based Point Cloud Compression} \label{sec:lr-pcc}
\begin{figure}
    \centering
    \includegraphics[width=0.9\linewidth]{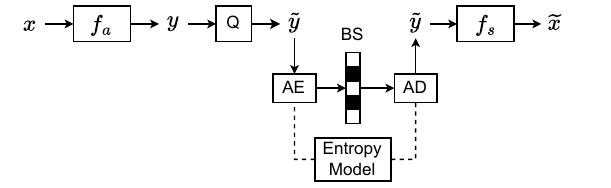}
    \caption{Autoencoder-style paradigm for learning-based compression. Q denotes quantization. AE denotes arithmetic encoder and AD denotes arithmetic decoder. BS denotes the binary bitstream generated by the arithmetic coder.}
    \label{fig:ae}
\end{figure}

Handcrafted compression methods rely on manually designed, limited context patterns and require additional bitrates to encode pattern selection, thus restricting compression performance. In contrast, learning-based compression methods automatically model data structure by traversing the entire dataset. 
% They use nonlinear transformations to map data from the original spatial domain to a compact latent feature representation, reducing data volume while retaining key information. 
These methods generally follow an autoencoder-style unified paradigm shown in Fig. \ref{fig:ae}. Specifically, a nonlinear analysis transform $f_a$ maps the input data $x$ to a compact latent feature $y$, reducing data volume while retaining key information. $y$ is next quantized and compressed in the bottleneck, and then a nonlinear synthesis transform $f_s$ maps the decompressed feature $\tilde{y}$ back to the spatial domain as a reconstruction of $x$. 
To achieve superior compression performance and enable end-to-end optimization for specified rate-distortion tradeoffs, entropy coding of the quantized latent feature $\tilde{y}$ is often modeled within the compression network additive uniform noise and \cite{pcc_balle2016end} typically replaces the non-differentiable quantization during training. The probability distribution of $\tilde{y}$ is usually estimated under a fully factorized assumption \cite{pcc_balle2016end} with side information like Gaussian scale hyperprior \cite{pcc_balle2018variational} and spatial priors\cite{pcc_minnen2018joint, pcc_he2021checkerboard}. 

Existing learning-based point cloud compression methods can be categorized into point-based, voxel-based, tree-based, and heterogeneity-based. We detail them as follows.

\subsubsection{Point-based Point Cloud Compression}
Point-based PC compression methods directly process discrete points in 3D using multilayer perceptron (MLP)-based autoencoders. They usually use PointNet\cite{pcc_qi2017pointnet} and PointNet++\cite{pcc_qi2017pointnet++} as feature extractors. PointNet is used to extract global features from 3D coordinates via MLPs and max-pooling while PointNet++ captures local structure features via farthest point sampling (FPS) and K-nearest neighbor (KNN) search. Huang \textit{et al.} \cite{pcc_huang20193d} employ hierarchical PointNet++ and multi-scale reconstruction loss for geometry compression. Later, Yan \textit{et al.} \cite{pcc_yan2019deep} and You \textit{et al.} \cite{pcc_you2021patch} introduce learnable entropy coding. NGS \cite{pcc_gao2021point} further proposes local feature aggregation before down-sampling and attention-based sampling with an MLP-based hyper-prior entropy model. For attribute compression, DeepPCAC\cite{pcc_sheng2021deep} designs second-order point convolutional layers and dense point-inception blocks to enlarge the receptive field.

The major weakness of point-based algorithms is their high computational and time complexity due to FPS and KNN, which limits their application to smaller datasets, such as ShapeNet\cite{pcc_chang2015shapenet} and ModelNet\cite{pcc_modelnet}. Additionally, point-based geometry coding neglects low-resolution coordinate information in the autoencoder bottleneck, either ignoring it\cite{pcc_huang20193d, pcc_yan2019deep} or simply concatenating it with latent features before compression\cite{pcc_you2021patch, pcc_gao2021point}. Moreover, MLP-based entropy models fail to capture spatial correlations. In geometry reconstruction, simply utilizing MLP to up-sample latent features and map them to 3D reconstruction coordinates is ineffective for dense point clouds. The above limitations affect the rate distortion performance and subjective quality of point-based compression methods.

\subsubsection{Voxel-based Point Cloud Compression}
Voxel-based methods partition raw data into 3D regular voxels like pixels in the 2D space, and utilize convolution in the 3D space to construct autoencoder networks. Quach \textit{et al.}\cite{pcc_quach2019learning} proposes the first lossy point cloud geometry compression framework based on 3D convolution layers, which models reconstruction as a binary classification task for voxel occupancy. Benefiting from the neighborhood feature aggregation capability of 3D convolution, subsequent works\cite{pcc_quach2020improved, pcc_pcgcv1} outperform MPEG G-PCC. However, their memory and computational complexity increase cubically with bit depth. Consequently, point clouds need to be divided into small cubes, which destroys the surface continuity and information flows between blocks.

To solve the problem, Wang \textit{et al.} propose PCGCv2\cite{pcc_pcgcv2} based on \cite{pcc_pcgcv1}, which replaces 3D convolutions with efficient sparse convolutions \cite{pcc_choy20194d} for lossy geometry compression. It represents the voxelized input point cloud as a sparse tensor, enabling calculations only for occupied voxels. Sparse convolutions manage and reuse mappings between the input and output coordinate sets to further reduce complexity, allowing deeper networks and full-point-cloud processing without block limitations. A top-K pruning strategy\cite{pcc_pcgcv1} is also introduced for hierarchical lossy reconstruction, which estimates occupancy probabilities via sparse convolutions and sigmoid activation, and only retains points with K-highest occupancy probabilities at each resolution. In the autoencoder bottleneck, latent features are lossily coded via a factorized entropy model\cite{pcc_balle2018variational}, while the associated occupancy map is losslessly encoded by G-PCC. PCGCv2 achieves a 70\% BD-Rate gain over G-PCC with a similar encoding time, and most existing voxel-based compression frameworks follow its architecture:

% \begin{figure}
%     \centering
%     \includegraphics[width=0.9\linewidth]{figures/pcgcv2.pdf}
%     \caption{Architecture of PCGCv2\cite{pcc_pcgcv2}. \emph{Sconv(C,K,S)} denotes a sparse convolution layer with output channels \emph{C}, kernel size \emph{K} and stride \emph{S}. \emph{TSconv} denotes a sparse transpose convolution layer. \emph{Q} denotes quantization.}
%     \label{fig:pcgcv2}
% \end{figure}

\textbf{Static Point Cloud Geometry Compression.} Building on PCGCv2\cite{pcc_pcgcv2}, Wang \textit{et al.} further propose SparsePCGC\cite{pcc_sparsepcgcv1} that integrates lossless and lossy geometry compression of dense objects and sparse LiDAR point clouds within a unified framework. This method achieves breakthrough performance that surpasses other handcrafted and learning-based static point cloud geometry compression methods, especially on dense objects. Its success can be attributed to three key innovations: an 8-stage SparseCNN-based occupancy probability approximation (SOPA) for sibling-referenced lossless compression, adjustable lossy coding layers for broader bitrate ranges, and a super-resolution module for maintaining high quality at extremely low bitrates. 
% First, it designs an efficient lossless compression module named 8-stage SparseCNN-based Occupancy Probability Approximation, which refers to coded sibling nodes within the same parent node. Second, the bit depth level number can be adjusted for lossy/lossless coding. Compared to fixing the number of lossy reconstruction layers and merely adjusting the RD weight in the loss function for rate control, this approach enables a broader bitrate range and better RD performance. Third, it incorporates a super-resolution module designed for dense object point clouds. At extremely low bitrates in lossy coding (\textless 0.05bpp), it can still maintain high objective quality (D1-PSNR\textgreater 70dB for 10-bit point clouds). 
However, SparsePCGC struggles to extract features from extremely sparse data. Subsequent methods like SparsePCGCv2\cite{pcc_sparsepcgcv2} and Unicorn\cite{pcc_unicorn1} replace sparse convolutions with transformers at higher bit depth levels, while Latent-Guided\cite{pcc_fan2023multiscale} maps the occupancy status of each bit depth level to a latent feature embedding for auxiliary guidance.

\textbf{Dynamic point cloud geometry compression.} Like video inter-frame coding, dynamic point cloud compression performs inter-prediction on time-consecutive frames. Some methods directly map latent features of previously reconstructed frames into temporal priors without explicit motion estimation (ME) and motion compensation (MC). For example, Akhtar \textit{et al.}\cite{pcc_akhtar} fuses multi-scale latent features of one previous frame as a prediction of the current frame via convolution on target coordinates at the autoencoder bottleneck, and then compress the latent feature residual between the prediction and the current frame via a factorized entropy model. \cite{pcc_m62066, pcc_m65094} further incorporate dual reference frames and cross-attention-like feature transformation. SparsePCGCv3\cite{pcc_sparsepcgcv3} and Unicorn \cite{pcc_unicorn1} replace residual coding with conditional coding for better redundancy removal. 
% As residual coding only uses the subtraction operation to remove the redundancy across frames, conditional coding directly codes latent features of the current frame with probability distribution parameters estimated by spatiotemporal priors and, thus, has a higher compression ratio\cite{pcc_dcvc}. 
D-DPCC\cite{pcc_tingyu-ddpcc} is the first end-to-end dynamic point cloud geometry compression framework with explicit ME and MC in the latent feature domain for dense objects. It proposes SparseCNN-based two-scale motion fusion for ME and distance-weighted adaptive interpolation for MC. LDPCC\cite{pcc_xia2023learning} and U-Motion\cite{pcc_u-motion} further refine this approach with multi-layer ME and MC architectures for coarse-to-fine inter-prediction.

\textbf{Attribute compression.} Most voxel-based attribute compression methods construct SparseCNN-based autoencoder frameworks like geometry compression. For lossy coding, LossyPCAC\cite{pcc_wang2022sparse} uses an auto-regressive-based entropy model, and TSC-PCAC\cite{pcc_TSC-PCAC} designs a transformer-based channel context entropy model. YOGA\cite{pcc_zhang2023yoga} utilizes G-PCC lossy coded color at the autoencoder bottleneck and concatenates it with latent features for decoding. Unicorn\cite{pcc_unicorn2} progressively compresses the attribute residuals between the voxels and their parent nodes to eliminate spatial redundancy and uses conditional coding in dynamic compression. U-Motion uses multi-layer ME and MC like in geometry coding. For lossless coding, CNeT\cite{pcc_cnet} performs auto-regressive reference within and between channels. CrossPCAC~\cite{pcc_wang2023lossless} and Unicorn~\cite{pcc_unicorn2} adopt the 8-stage SOPA from SparsePCGC and update attribute residuals by sibling voxels. Some works \cite{pcc_NF-PCAC, pcc_lin2023sparse} explore normalized flow structures with strictly reversible modules rather than autoencoders, but they are less efficient compared to autoencoder-style methods.

\subsubsection{Tree-based Point Cloud Compression}
Tree-based methods represent the original point cloud with a tree structure and then process tree nodes with transformer-style networks. These methods are particularly suitable for sparse point clouds such as LiDAR data, where the octree structure is most commonly used because it is easily constructed and compatible with convolution operations. 
Specifically, an octree with a maximum depth $L$, $\textbf{x} = \{x^{(1)}, x^{(2)}, \ldots, x^{(L)} \}$, can be built for a point cloud with geometry bit depth $L$. 
% From top to down, an 8-bit occupancy code is assigned for each node to indicate its eight child nodes' occupancy status, so octree construction and reconstruction is always modeled as a 255-dimensional multi-classification task. 
Similar to handcrafted octree-based point cloud compression, the occupancy status distribution can be predicted auto-regressively:
\begin{equation}\label{octree}
q(\textbf{x}) = \prod\limits_{i}^{N} q\left( x_{i} \mid context(x_{1:i-1}) \right)
\end{equation}
where $q(\textbf{x})$ denotes the estimated occupancy probability distribution of all points in $\textbf{x}$; $N$ is the number of points in $\textbf{x}$; $context(x_{1:i-1})$ denotes a context set derived from all nodes decoded before the current node $x_{i}$ or a subset of them.

The strategy of determining the context set varies among methods. Initially, OctSqueeze\cite{pcc_huang2020octsqueeze} defines the context of a given node as its parent and ancestor nodes, as shown in Eq.~\ref{octsqueeze}, where $x^{(l)}$ denotes the occupancy status of nodes at level $l$ of the octree. The method assumes the independence of the nodes at the same octree level, enabling top-to-bottom encoding and decoding with low complexity and good parallelism. However, ignoring strong spatial correlations among sibling nodes causes suboptimal distribution estimation and inefficient compression.
\begin{equation}\label{octsqueeze}
q(\textbf{x}) = \prod\limits_{l=1}^{L} q\left( x^{(l)} \mid x^{(1:l-1)} \right)
\end{equation}
To address this problem, VCN\cite{pcc_que2021voxelcontext} utilizes 3D convolution networks to capture local dependencies among same-level nodes in bottom octree levels. OctAttention\cite{pcc_fu2022octattention} expands the ancestor and sibling contexts with a serial context window updating mechanism: It traverses each octree node with a breadth-first search (BFS) order, maintaining a context window that stores ancestors, siblings, and their ancestors. The contexts pass through a self-attention block for occupancy estimation, as shown in Eq. \ref{octattention}: 
\begin{equation}\label{octattention}
q(x^{(l)} \mid x^{(1:l-1)}) = \prod\limits_{i=1}^{N_l} q\left( x^{(l)}_i \mid x^{(1:l-1)}, sib(x^{(l)}_i) \right)
\end{equation}
where $x^{(l)}_i$ denotes the $i$-th node at level $l$. $N_l$ denotes the number of points at level $l$ and $ sib(x^{(l)}_i)$ are coded sibling nodes of the current node $x^{(l)}_i$. However, OctAttention requires updating the context window and re-running the network for decoding each node, resulting in non-parallelism.

To balance complexity and compression efficiency, emerging work optimizes feature extraction modules and designs various grouping strategies. ECM-OPCC \cite{pcc_ecm-opcc} propose a dual transformer module and multi-group coding strategy. EHEM \cite{pcc_song2023efficient} design a hierarchical attention model and introduce the checkerboard strategy \cite{pcc_he2021checkerboard}, achieving state-of-the-art performance on static LiDAR point cloud geometry compression. SCP \cite{pcc_scp} construct octrees in the spherical coordinate system for LiDAR point clouds and outperform methods in the Cartesian coordinate system by up to 29\% BD-Rate gain.
% \begin{equation}\label{ftylidar}
% q(x^{(l)} \mid x^{(1:l-1)}, f^{(l)}) = \prod\limits_{i=1}^{N_l} q\left( x^{(l)}_i \mid x^{(1:l-1)}, f^{(l)}) \right)
% \end{equation}

\subsubsection{Heterogeneity-based Point Cloud Compression}
Heterogeneous methods unify various point cloud representations to compress an input point cloud at different bit depth levels. First and foremost, it should be noted that most learning-based point cloud compression networks rely on tree-based handcrafted coders to compress the lowest bit depth information at the autoencoder bottleneck. For example, voxel-based geometry coding\cite{pcc_pcgcv2, pcc_sparsepcgcv2, pcc_unicorn1, pcc_akhtar, pcc_m62066, pcc_m65094, pcc_sparsepcgcv3, pcc_tingyu-ddpcc, pcc_xia2023learning, pcc_u-motion} uses G-PCC octree for lossless coding of the lowest bit coordinates. Voxel-based attribute coding \cite{pcc_zhang2023yoga, pcc_unicorn2, pcc_wang2023lossless} uses G-PCC RAHT for lossy coding of the lowest bit attributes. Some point-based geometry coding methods \cite{pcc_you2021patch} use G-PCC octree for lossless coding of partition information at certain octree levels. Since these bits are handled without learning-based algorithms, such methods are not considered heterogeneous and are classified as purely voxel-based or point-based. Heterogeneity-based point cloud compression only concerns the remaining bits. 

\begin{figure}
    \centering
    \includegraphics[width=0.9\linewidth]{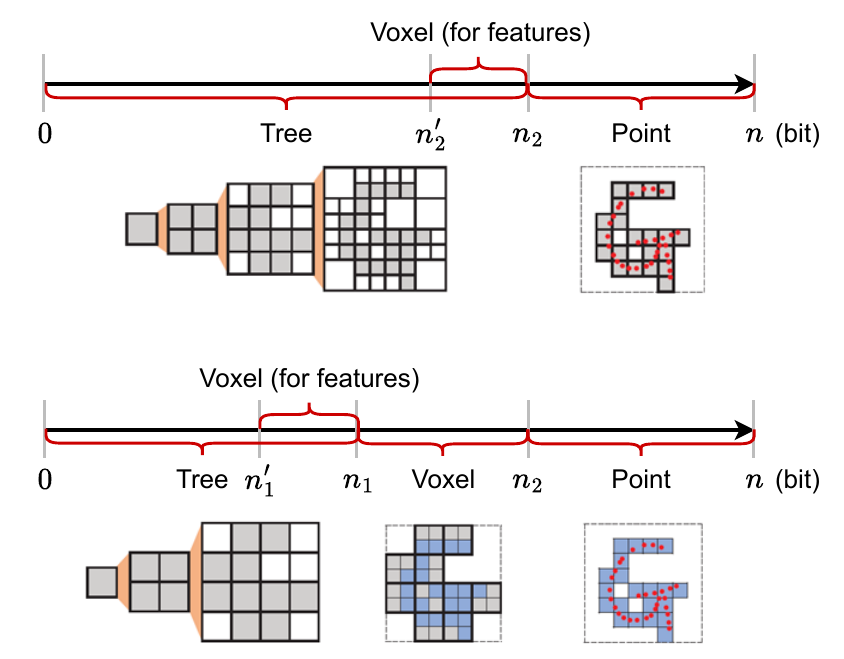}
    \caption{Framework of GRASP-Net (top) and PIVOT-Net (bottom).}
    \label{fig:graspnet}
    \vspace{-0.5cm}
\end{figure}

GRASP-Net\cite{pcc_pang2022grasp} and PIVOT-Net\cite{pcc_pang2024pivot} are the first heterogeneity-based point cloud compression frameworks: Aimed at geometry coding, they incorporate SparseCNN layers into a PointNet++ based autoencoder architecture for entropy coding and latent feature resolution reduction. For a point cloud with bit depth $n$, GRASP-Net handles a bit depth interval $[n^\prime_2, n_2]$, where 0\textless $n^\prime_2$\textless $n_2$\textless $n$. It first generates a coarse version of the original point cloud by uniform quantization with a step $s=2^{n-n_2}$. Then, a PointNet++ based analysis block encapsulates fine geometric details of the highest $n-n_2$ bits into latent feature embeddings, and the $n_2$-bit coarse version is losslessly coded by G-PCC octree. SparseCNN downsamples latent features associated with $n_2$-bit coordinates to $n^\prime_2$-bit for factorized entropy coding. On the decoder side, the $n_2$-bit coordinates are decoded while the latent features are upsampled, and a PointNet++ synthesis block maps features to 3D coordinate offsets for reconstruction.
By combining the larger receptive field of point-based blocks at higher bit depths with the neighborhood correlation capture of voxel-based blocks at lower bit depths, GRASP-Net outperforms voxel-based frameworks like SparsePCGC on sparse surface and LiDAR datasets. PIVOT-Net further splits the tree-based bit interval, as depicted in Fig. \ref{fig:graspnet}. Geometric information between $n_2$-bit and $n_1$-bit is extracted by SparseCNN and reconstructed with Sparse Transpose CNN and binary classification as in \cite{pcc_pcgcv2, pcc_sparsepcgcv1}. Thus, PIVOT-Net can surpass SparsePCGC even in solid surface and dense surface point clouds by about 10\% BD-Rate gain.

Given the advantages of heterogeneous methods, MPEG AI-based point cloud coding reference software, TMAP\cite{pcc_tmapv1}, also adopts a heterogeneous geometry coding framework. For sparse static and dynamic LiDAR point clouds, TMAP adopts a structure similar to PIVOT-Net. For dense static and dynamic objects, it employs a purely voxel-based structure similar to Unicorn\cite{pcc_unicorn1, pcc_unicorn2}. TMAP’s architecture is also extended to attribute coding by replacing the 3D point geometry with attribute information like color and reflectance, known as UniFHiD~\cite{pcc_unifhid-part3}.

\subsection{Performance Evaluation}
\subsubsection{Experiment Settings}
\ 
\newline
\indent \textbf{Database.} To fairly compare point cloud compression methods, we select test datasets following the Common Test Condition (CTC) on MPEG AI-based point cloud coding (AI-PCC) \cite{pcc_ctc}, which divides point clouds into four categories: 
\begin{itemize}
    \item Category 1 Dense Static (C1 RWTT), sampled from textured real-world 3D meshes in the RWTT (Real World Textured Things) dataset \cite{pcc_RWTT} for immersive applications.
    \item Category 2 Sparse Static (C2 Static), sparse voxelized point cloud frames representing non-human object and scene surfaces from museum scans for immersive applications.
    \item Category 3 Dense Dynamic (C3 Dynamic), sequences of dynamic human bodies scanned from real humans for immersive applications.
    \item Category 4 Sparse Dynamic (C4 LiDAR), sequences of LiDAR scans from the KITTI dataset \cite{pcc_KITTI} obtained in outdoor self-driving, with raw LiDAR data and other data modalities for autonomous navigation and robotics.
\end{itemize}

For training datasets of learning-based algorithms, methods evaluated on C3 and C4 all follow the TrainVal sets specified in MPEG AI-PCC CTC. Methods for C1 and C2 use different training materials: PCGCv2~\cite{pcc_pcgcv2} and SparsePCGC~\cite{pcc_sparsepcgcv1} use ShapeNet~\cite{pcc_chang2015shapenet}, GRASP-Net~\cite{pcc_pang2022grasp} uses ModelNet40~\cite{pcc_modelnet}, Unicorn~\cite{pcc_unicornv2geo, pcc_unicornv2attr} and TMAP~\cite{pcc_tmapv1} follow CTC. All learning-based methods follow the original authors' data preprocessing and augmentation settings to ensure optimal performance.

\textbf{Test Conditions.} We follow the three test conditions defined in MPEG AI-PCC CTC:
\begin{itemize}
    \item Track 1 (T1): Lossy Geometry
    \item Track 2 (T2): Lossy Geometry + Lossy Attribute
    \item Track 3 (T3): Lossless Geometry
\end{itemize}

\textbf{Evaluation Criteria.} We use bits per input point (bpip) to evaluate the bitrates for both lossy and lossless point cloud compression, D1 (point-to-point) PSNR and D2 (point-to-plane) PSNR to evaluate the distortion in lossy compression. For dynamic sequences, bpip is calculated as the total bitstream size divided by the total number of points across all frames. PSNR values are obtained by averaging across all frames. We use MPEG-3DG \emph{mpeg-pcc-dmetric} metric software version 0.13.5 \cite{pcc_pcerror_software} to calculate D1 and D2 PSNR.

For C1 to C3, geometry PSNR peak values are set to $2^{geo\_bitdepth}-1$, where $geo\_bitdepth$ is the bit depth of coordinates. Color attribute PSNR peak value is 255, as the data have 8-bit color attributes. For C4, original floating-point coordinates are first divided by 0.001 and then rounded to integers in the range $[-2^{17}, 2^{17}-1]$. Geometry PSNR peak value is set to 30000, and reflectance PSNR peak value is 65535 since each point has 16-bit reflectance attributes.
% BD-Rate and CR?

\textbf{Handcrafted baselines.} We use three MPEG-standardized open source software to evaluate handcrafted point cloud compression methods: TMC13-v28.0 \cite{pcc_gpcc_software}, GeSTM-v9.0\cite{pcc_gestm_software} and TMC2-v24.0 \cite{pcc_vpcc_software}. Specifically, we utilize the following configurations:
\begin{itemize}
    \item TMC13-v28.0 (marked as G-PCC):
        \begin{itemize}
        \item Lossy and Lossless geometry coding: Octree
        \item Lossy attribute coding: RAHT
        \item Modes: Intra-frame
        \end{itemize}
    \item GeSTM-v9.0 (marked as GeSTM):
        \begin{itemize}
        \item Lossy geometry coding: Trisoup
        \item Lossless geometry coding: Octree
        \item Lossy attribute coding: RAHT
        \item Modes: Intra-frame and inter-frame
        \end{itemize}
    \item TMC2-v24.0 (marked as V-PCC):
        \begin{itemize}
        \item Video codec: HM % HEVC Test Model
        \item Modes:
            \begin{itemize}
            \item Lossy: All Intra (AI), Random Access (RA) % with Group of Frame size 32
            \item Lossless: Low Delay (LD)
            \end{itemize}
        \end{itemize}
\end{itemize}

\subsubsection{Compression of Dense Point Cloud}

\begin{figure}
    \centering
    \includegraphics[width=0.9\linewidth]{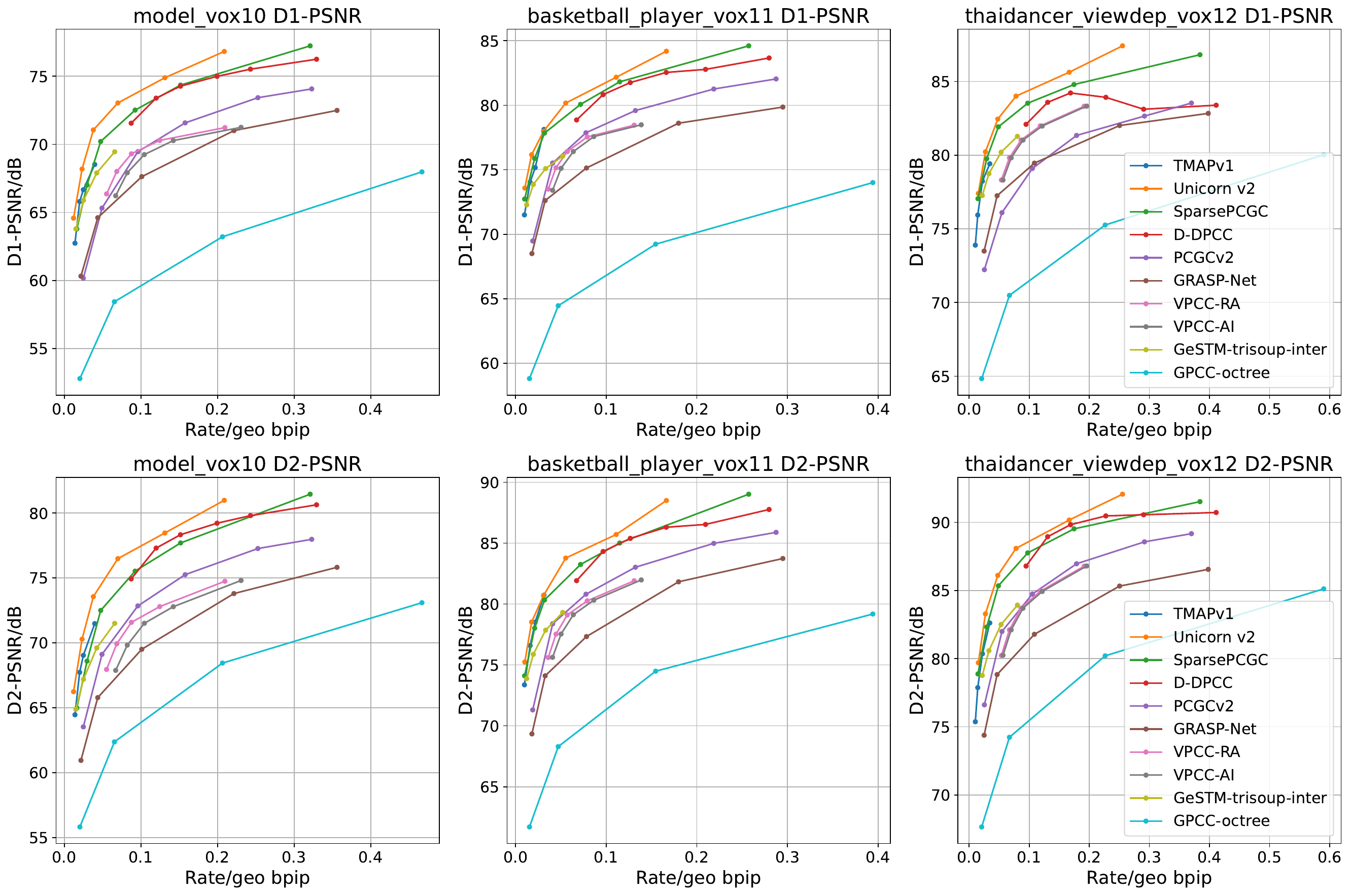}
    \caption{Lossy-geometry results on dense dynamic datasets.}
    \label{fig:T1_dense_dynamic}
    \vspace{-0.3cm}
\end{figure}

\begin{figure}
    \centering
    \includegraphics[width=0.9\linewidth]{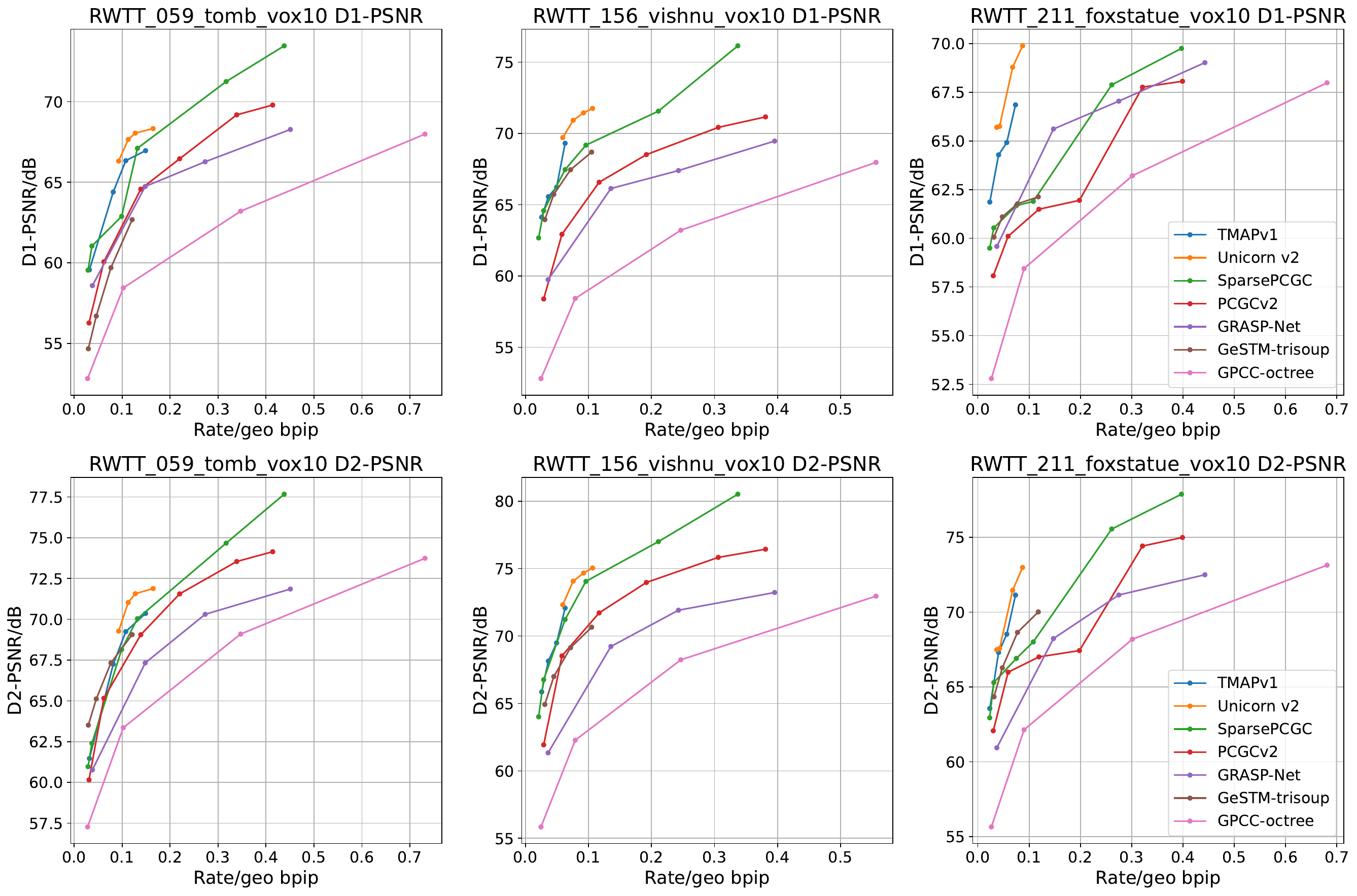}
    \caption{Lossy-geometry results on dense static datasets.}
    \label{fig:T1_dense_static}
    \vspace{-0.4cm}
\end{figure}
\ 
\newline
\indent\textbf{Geometry Compression of Dense Point Cloud.} Fig.~\ref{fig:T1_dense_dynamic} and Fig.~\ref{fig:T1_dense_static} present a comprehensive evaluation of lossy geometry compression performance in dense point clouds.

For handcrafted methods, V-PCC and GeSTM outperform G-PCC octree and even demonstrate comparable performance over learning-based methods because theybenefit from mature video compression codecs and adaptive surface approximation capabilities, respectively. For learning-based methods, those integrating 8-stage SOPA in lossless coding, along with variable lossy coding layers (TMAPv1\cite{pcc_tmapv1}, Unicorn v2\cite{pcc_unicornv2geo}, and SparsePCGC), outperform those methods maintaining a fixed number of lossy coding layers and merely adjusting the RD weight or pre-quantization step for rate control, such as D-DPCC, PCGCv2 and GRASP-Net. Additionally, while both D-DPCC and TMAPv1 incorporate explicit motion estimation and compensation for dynamic sequences, an intriguing phenomenon is observed in D-DPCC where an increase in bitrate correlates with a degradation in D1-PSNR for 12-bit data. This undesirable effect does not exist in TMAPv1, which is also attributed to its implementation of variable lossy layers.

Lossless geometry: Methods supporting lossless geometry coding and their bpip results are presented in Tabs. \ref{tab:T3_dense_dynamic} and \ref{tab:T3_dense_static}. 
They perform similarly to lossy coding, except for V-PCC, which exhibits limited performance compared to the G-PCC octree due to its constraints in preserving exact geometric information efficiently.
% Unlike in the lossy scenario, TMAPv1 and Unicorn v2 are outperformed by SparsePCGC on the \emph{thaidancer} dataset, despite incorporating inter-frame prediction, which the latter lacks. This derives from training randomness. 
% SparsePCGC trained by 8i+queen has similar lossless results to Unicorn and TMAP, trained by shapenet performs extremely better on thaidancer. 

\begin{table*}[tbp]
    \caption{Lossless-geometry results on dense dynamic datasets.}
    \label{tab:T3_dense_dynamic}
    \begin{center}
        \begin{tabular}{l|ccc}
        \toprule
        {} & \textbf{model\_vox10} & \textbf{basketball\_player\_vox11} & \textbf{thaidancer\_viewdep\_vox12} \\ 
        \hline
        \textbf{TMAPv1} & 0.598 & 0.452 & 1.253 \\
        \textbf{Unicorn v2} & 0.597 & 0.450 & 1.536 \\
        \textbf{SparsePCGC} & 0.660 & 0.494 & 0.973 \\
        \textbf{V-PCC (LD)} & 1.460 & 1.093 & 1.490 \\
        \textbf{GeSTM (octree-inter)} & 0.636 & 0.475 & 1.472\\
        \textbf{G-PCC (octree)} & 0.750 & 0.617 & 0.840 \\
        \bottomrule
    \end{tabular}
    \end{center}
\end{table*}

\begin{table*}[tbp]
    \caption{Lossless-geometry results on dense static datasets.}
    \label{tab:T3_dense_static}
    \begin{center}
        \begin{tabular}{l|ccc}
        \toprule
        {} & \textbf{RWTT\_059\_tomb\_vox10} & \textbf{RWTT\_156\_vishnu\_vox10} & \textbf{RWTT\_211\_foxstatue\_vox10} \\
        \hline
        \textbf{TMAPv1} & 1.165 & 0.914 & 1.019 \\
        \textbf{Unicorn v2} & 1.221 & 0.877 & 0.987 \\
        \textbf{SparsePCGC} & 1.125 & 0.913 & 1.061 \\
        \textbf{GeSTM (octree)} & 1.216 & 0.741 & 1.012 \\
        \textbf{G-PCC (octree)} & 1.235 & 0.766 & 1.058 \\
        \bottomrule
    \end{tabular}
    \end{center}
\end{table*}

\begin{figure}
    \centering
    \includegraphics[width=0.9\linewidth]{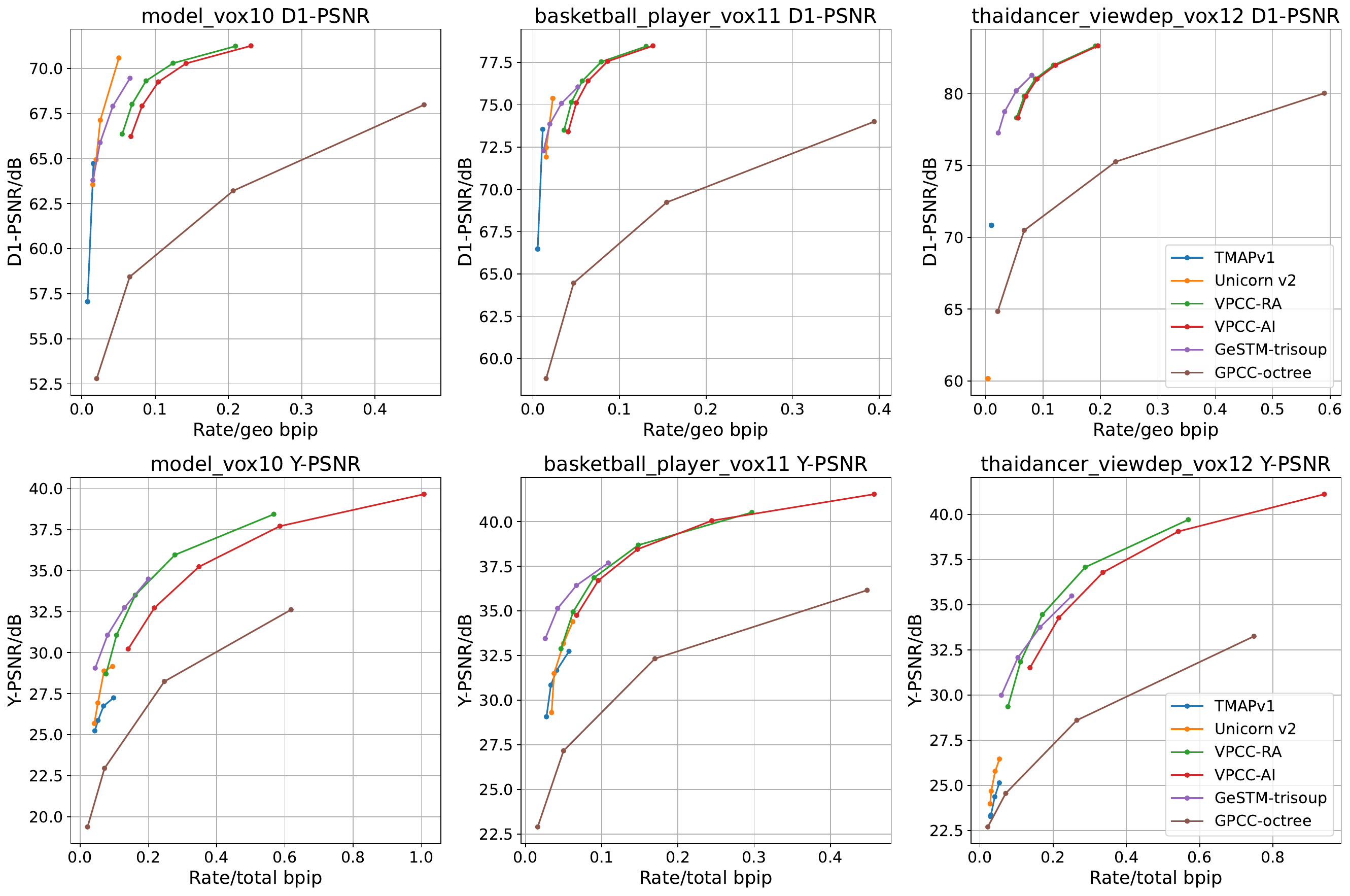}
    \caption{Lossy-geometry-lossy-attribute results on dense dynamic datasets.}
    \label{fig:T2_dense_dynamic}
    \vspace{-0.3cm}
\end{figure}

\begin{figure}
    \centering
    \includegraphics[width=0.9\linewidth]{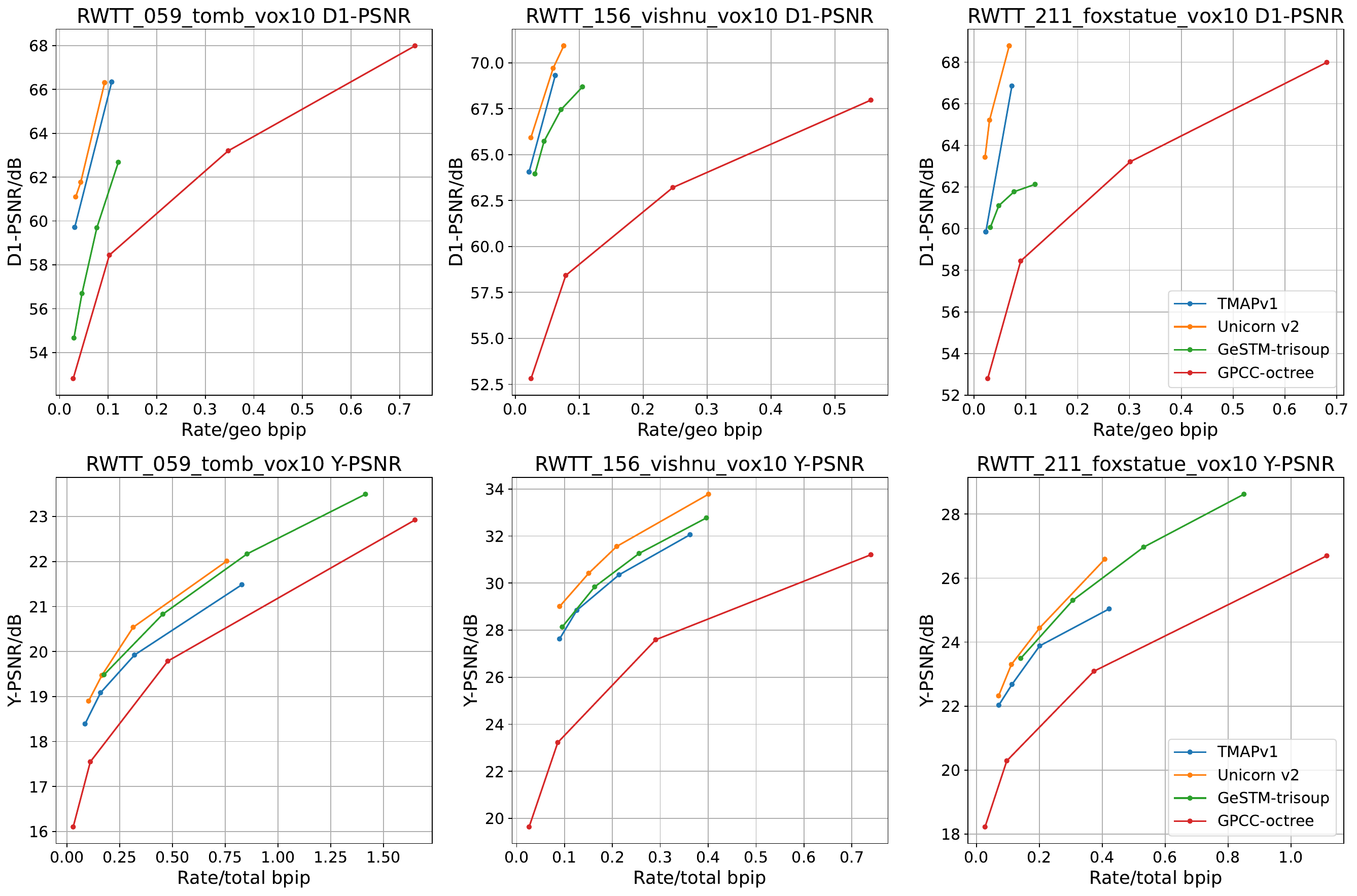}
    \caption{Lossy-geometry-lossy-attribute results on dense static datasets.}
    \label{fig:T2_dense_static}
    \vspace{-0.5cm}
\end{figure}

\indent\textbf{Attribute Compression of Dense Point Cloud.}
Figs.~\ref{fig:T2_dense_dynamic} and \ref{fig:T2_dense_static} illustrate lossy attribute compression performance on dense point clouds with lossy geometry. Unicorn v2 demonstrates superior performance in compressing attributes of dense point clouds, even under worse geometric quality. This advantages proves the efficiency of its cross-scale attribute residual coding and implicit inter-prediction.
% Future work could focus on integrating the superior performance of Unicorn v2 in static scenes with the benefits of UniFHiD in dynamic low-bitrate scenarios to create a balanced solution. 

We provide the following design insights for the compression of dense point clouds:
\begin{itemize}
\item Adaptive geometric representation. Inspired by the performance disparity between octree- and trisoup-based handcrafted methods, future learning-based compression frameworks should consider combining the strengths of voxel-based and surface-based approaches, especially for dense objects with uniform density and closed surfaces.
\item Multi-granularity rate control. We observe significant performance improvements of variable lossy layers over fixed layer coding structures. This suggests more granular rate control mechanisms, which could include dynamic layer allocation, adaptive RD weights and quantization steps that can be optimized across different regions within a single point cloud frame and across consecutive frames in dynamic sequences.
\item Hybrid feature extraction architectures. Although voxel-based learning methods currently outperform heterogeneous architectures like GRASP-Net in dense data, there remains potential for the combination of complementary feature extraction strategies. Future methods could fuse the spatial regularity of voxel-based operations and the geometric sensitivity of point-based modules.
\end{itemize}

\subsubsection{Compression of Sparse Point Cloud}
\begin{figure*}[htb]
    \centering
    \includegraphics[width=\linewidth]{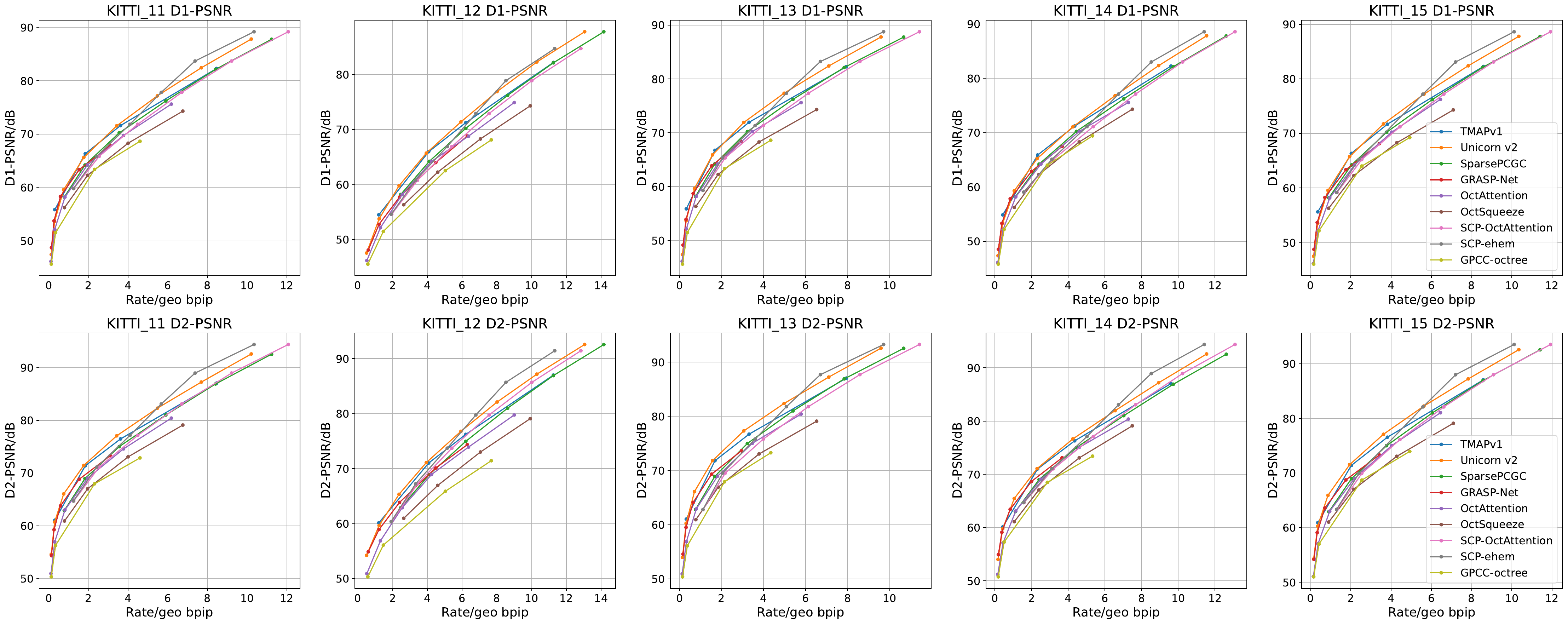}
    \caption{Lossy-geometry results on sparse dynamic datasets.}
    \label{fig:T1_sparse_dynamic}
\end{figure*}

\begin{figure*}[htb]
    \centering
    \includegraphics[width=\linewidth]{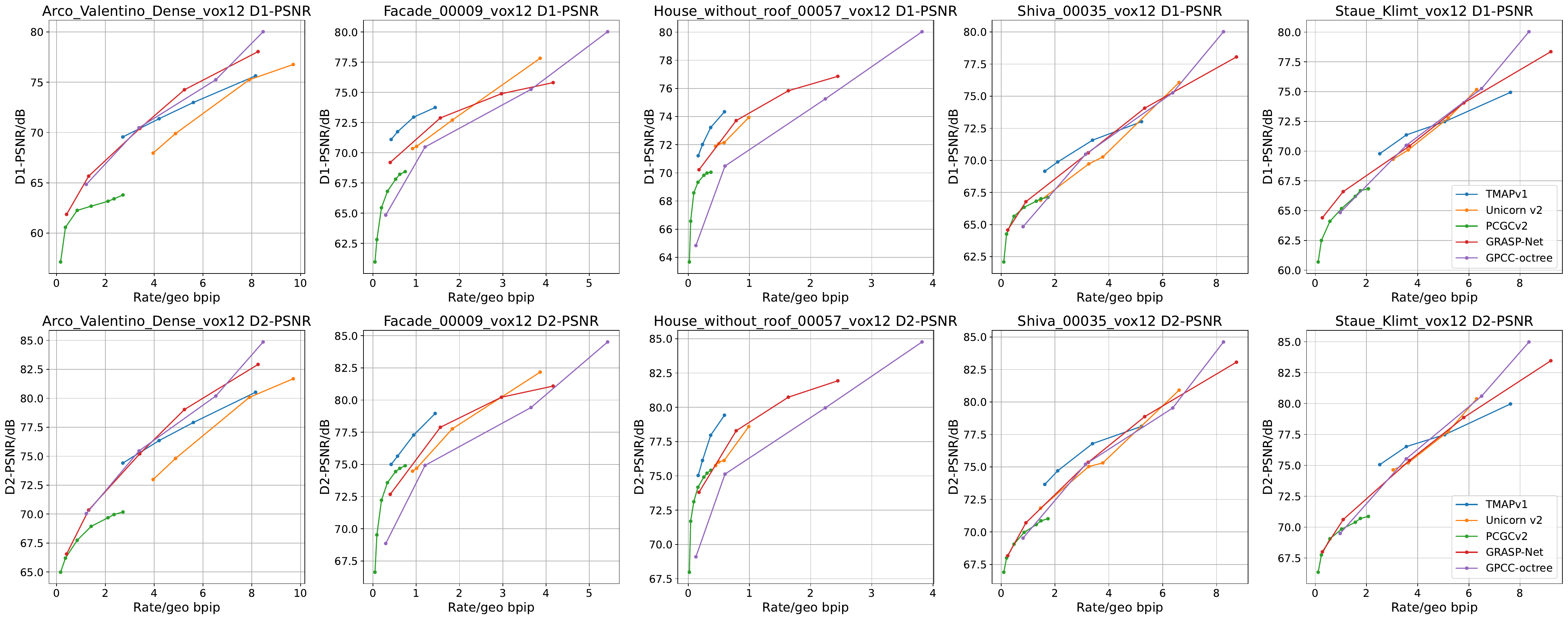}
    \caption{Lossy-geometry results on sparse static datasets.}
    \label{fig:T1_sparse_static}
\end{figure*}

\
\newline
\indent\textbf{Geometry Compression of Sparse Point Cloud.}
Detailed results of lossy geometry compression performance on sparse dynamic and static point clouds are provided in Figs. \ref{fig:T1_sparse_dynamic} and \ref{fig:T1_sparse_static}. 

For the sparse dynamic KITTI dataset, learning-based methods consistently outperform the G-PCC octree. 
All methods employ a global-quantization-based rate control strategy: It quantizes the original point cloud before entering a lossless compression network, and dequantize to restore the original scale. Thus, TMAPv1, Unicorn v2, and SparsePCGC, with 8-stage SOPA in lossless coding and coordinate refinement after inverse quantization, outperform simpler tree-based methods, such as OctSqueeze, OctAttention, and SCP-OctAttention\cite{pcc_scp}. Among them, Unicorn v2 excels with inter-frame prediction and transformer-based feature aggregation but is surpassed by SCP-ehem\cite{pcc_scp} at higher bitrates, due to both spherical coordinate-based neighborhood modeling and advanced entropy model design. For sparse static datasets, TMAPv1 outperforms all other methods on two denser point clouds, \emph{Facade} and \emph{House}, but is less efficient on others at high bitrates, requiring further optimization of the combination of the voxel-based and point-based modules.

\begin{table*}[tbp]
    \caption{Lossless-geometry results on sparse dynamic datasets.}
    \label{tab:T3_sparse_dynamic}
    \begin{center}
        \begin{tabular}{l|ccccc}
        \toprule
        {} & \textbf{KITTI\_11} & \textbf{KITTI\_12} & \textbf{KITTI\_13} & \textbf{KITTI\_14} & \textbf{KITTI\_15} \\ 
        \hline
        \textbf{TMAPv1} & 17.129 & 20.056 & 16.608 & 18.383 & 17.278 \\
        \textbf{Unicorn v2} & 15.474 & 17.915 & 14.813 & 16.712 & 15.688 \\
        \textbf{SparsePCGC} & 17.182 & 20.132 & 16.632 & 18.568 & 17.364 \\
        \textbf{G-PCC (octree)} & 21.423 & 24.813 & 21.309 & 21.987 & 21.762 \\
        \bottomrule
    \end{tabular}
    \end{center}
\end{table*}

\begin{table*}[tbp]
    \caption{Lossless-geometry results on sparse static datasets.}
    \label{tab:T3_sparse_static}
    \begin{center}
        \begin{tabular}{l|ccccc}
        \toprule
        {} & \textbf{Arco\_Valentino\_Dense\_vox12} & \textbf{Facade\_00009\_vox12} & \textbf{House\_without\_roof\_00057\_vox12} & \textbf{Shiva\_00035\_vox12} & \textbf{Staue\_Klimt\_vox12} \\
        \hline
        \textbf{TMAPv1} & 12.498 & 7.493 & 5.707 & 10.910 & 11.107 \\
        \textbf{Unicorn v2} & 9.958 & 6.095 & 4.789 & 9.083 & 9.051 \\
        \textbf{G-PCC (octree)} & 9.901 & 6.536 & 4.898 & 9.468 & 9.553 \\
        \bottomrule
    \end{tabular}
    \end{center}
\end{table*}

\begin{figure*}[htb]
    \centering
    \includegraphics[width=\linewidth]{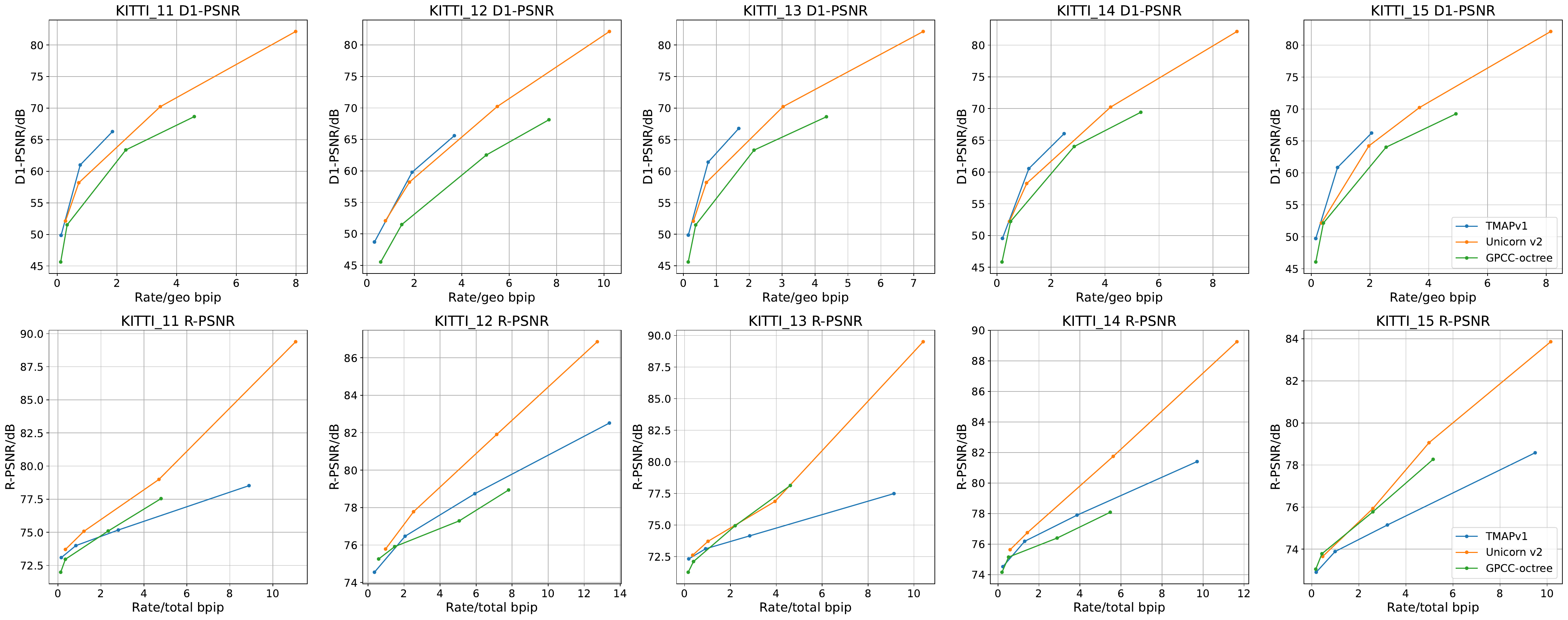}
    \caption{Lossy-geometry-lossy-attribute results on sparse dynamic datasets.}
    \label{fig:T2_sparse_dynamic}
\end{figure*}

\begin{figure*}[htb]
    \centering
    \includegraphics[width=\linewidth]{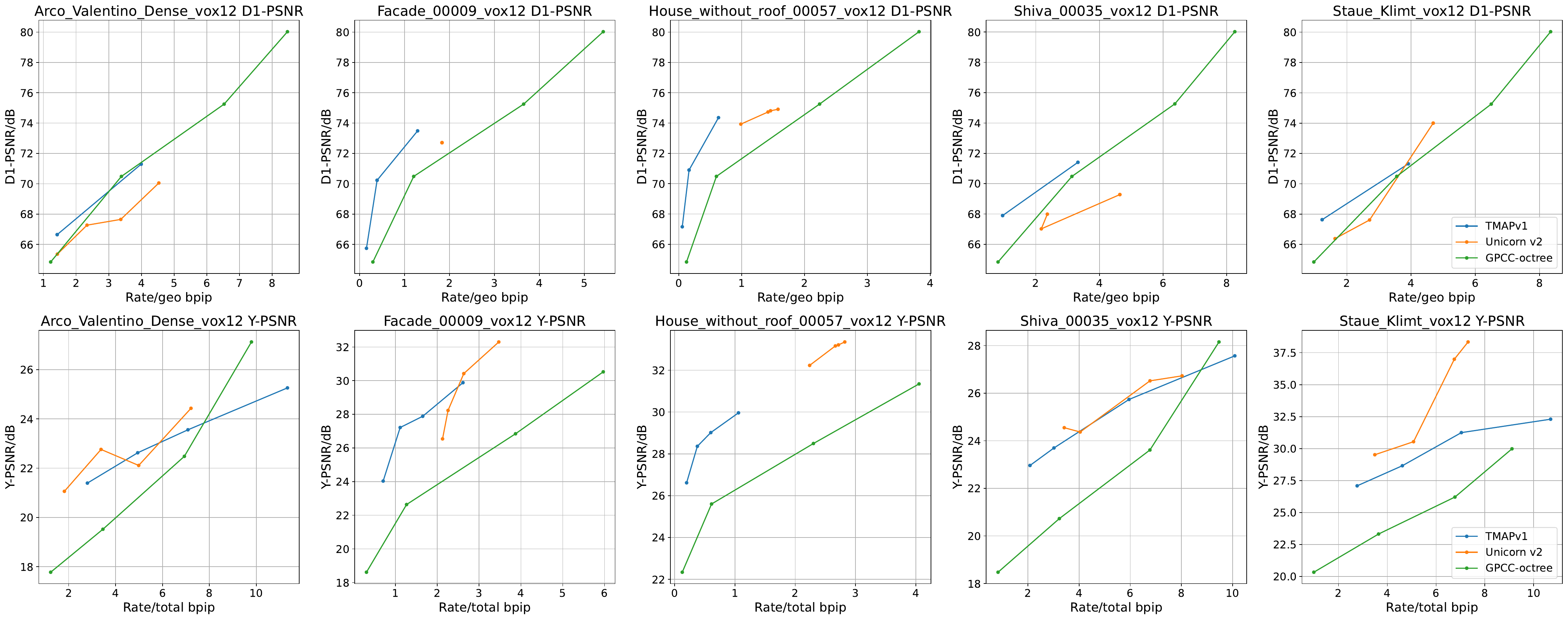}
    \caption{Lossy-geometry-lossy-attribute results on sparse static datasets.}
    \label{fig:T2_sparse_static}
\end{figure*}

\textbf{Attribute Compression of Sparse Point Cloud.}
The performance of lossy attribute compression on sparse point clouds with lossy geometry is shown in Figs. \ref{fig:T2_sparse_dynamic} and \ref{fig:T2_sparse_static}. Unicorn demonstrates superior attribute compression performance, even under conditions of suboptimal geometry reconstruction quality, but exhibits significant performance fluctuations on some datasets. In contrast, UniFHiD demonstrates relatively stable performance across different bitrates in static scenarios. This stability is attributed to its PointNet-based prediction, which is well-suited for the weaker spatial correlation of sparse data.

Currently, learning-based point cloud attribute compression methods do not significantly outperform handcrafted methods, perhaps because current attribute compression methods are migrated from geometric compression methods. Exploring alternative grouping methods, such as octree or kd-tree, could also provide more efficient compression strategies for the attribute compression of point clouds. In addition, future attribute compression methods should leverage decoded geometric information to guide attribute prediction, particularly for complex point clouds with varying density and attribute distributions.

We propose the following design insights for the compression of sparse point clouds:
\begin{itemize}
\item Dynamic quantization strategy. Sparse point clouds often exhibit irregular distributions and significant gaps. Instead of uniform quantization, context-aware mechanisms that adaptively allocate finer quantization steps to dense regions close to scanners can optimize rate-distortion performance.
\item Efficient temporal reference. For dynamic LiDAR sequences, existing methods utilize implicit temporal mapping or simple motion embedding without interpolation-based compensation. There are still challenges for efficient inter-frame block matching due to ego-motion, sparsity and non-uniformity of LiDAR data.
\item Advanced entropy models. The irregular distribution of sparse point clouds presents significant challenges for entropy coding. While existing methods leverage architectures like Swin-Transformer to extract meaningful features and model occupancy probability distributions, there is a critical need to balance local dependencies removal with storage efficiency and computational complexity.
\end{itemize}

\section{Point Cloud Quality Assessment}\label{sec:PCQA}

\subsection{Full-reference (FR) Point Cloud Quality Assessment}
\subsubsection{Model-based Metrics} The model-based metrics directly infer point cloud quality from its original 3D representation. Based on the minimum processing unit, these metrics can be categorized into point-based and structure-based metrics. Point-based metrics aim to gauge the absolute difference between the matched points, while structure-based metrics assume that structural change in local
regions (\textit{e.g.}, gradient \cite{pcqa_yang2020_graphsim}, curvature \cite{pcqa_meynet2020_pcqm}) dominates the quality decision, thus quantifying perceptual deformation at a macro level. 

\textbf{Point-based Metrics.} Early point-based metrics measure the distance between each point $x_i$ of the test point cloud $X$ and its nearest corresponding point $y_j$ in the reference point cloud $Y$.
For the error vector $\mathbf{e}(x_i, y_j)$, point-to-point (p2po) distance \cite{pcqa_p2p} is computed as the absolute 3D Euclidean distance:
\begin{equation}
    d_{X,Y}^{P2Po} = \| \mathbf{e}(x_i, y_j) \|_2
\end{equation}
Recognizing that point clouds represent coarse geometric approximations of surfaces, the point-to-plane (p2pl) metric \cite{pcqa_p2pl} refines this approach by projecting the p2po error along the normal direction $\mathbf{n_j}$ of the corresponding reference point:
\begin{equation}
    d_{X,Y}^{P2Pl} = \| \mathbf{e}(x_i, y_j) \cdot \mathbf{n_j} \|_2
\end{equation}
Then per-point errors are calculated for all the points and then aggregated using either the root mean square (RMS) distance (\( \text{p2po}_{\text{rms}} \)) or the Hausdorff distance (\( \text{p2po}_{\text{Haus}} \)).
Since the point correspondences between $T$ and $R$ are not identical, distances are typically computed in both directions, and the maximum value is taken as the final score.
% \begin{figure}[t]
%     \centering
%     \includegraphics[width=0.90\linewidth]{image/p2pl.pdf}
%     \caption{Paradigm of p2pl and p2po.}
%     \label{fig:p2p}
% \end{figure}
Given that color is also a critical factor in quality assessment, the RMS errors for the YUV attributes are also computed. Finally, the symmetric RMS error is normalized by the signal peak to compute the PSNR value. 
In MPEG \cite{pcqa_mpegCTC}, the p2po PSNR metric is called D1, the p2pl PSNR metric is called D2 and the averaged color PSNR metric is denoted as PSNR$_\text{YUV}$.

As point-based metrics have been adopted by MPEG as common test conditions, researchers have proposed various improvements to these metrics in recent years, mostly focusing on refining pooling and normalization strategies.
The classical Hausdorff distance metric, which considers only the maximum point error, is highly sensitive to a few outlier points in the distorted point cloud. To address this limitation, Javaheri \textit{et al.} \cite{pcqa_HausdorffGeneralized} propose a generalized Hausdorff distance metric that represents geometric quality considering a fixed portion of the ranked point errors. Javaheri \textit{et al.} 
\cite{pcqa_javaheri2020_improving} propose normalizing the MSE values using the estimated intrinsic resolution, which is determined as the average distance between the nearest points in 3D or 2D space, providing a more meaningful measure by accounting for the actual spatial distribution. Wang \textit{et al.} 
\cite{pcqa_pointNPM} introduce a method to calculate noticeable probability maps for each 3D point. By unprojecting a 2D just-noticeable difference map into 3D, these probabilities are used to weight the errors in point-based metrics.

\textbf{Structure-based Metrics.} Structure-based metrics inherit the idea of structural similarity (SSIM) and aim to measure macroscopic structural deformation. The early metrics mostly try to capture the changes in geometry structure. As a pioneering work, plane-to-plane \cite{alexiou2018point} utilizes the normal vector of two points to derive
the angular similarity. Instead of using normals, PC-MSDM \cite{meynet2019pc} uses local curvature statistics to measure structural distortion by fitting a quadric surface. MSMD \cite{javaheri2020mahalanobis} seizes the Mahalanobis distance between a point and a distribution, eliminating the need to calculate normals or fitting surfaces.

\begin{figure}[t]
    \centering
    \includegraphics[width=0.7\linewidth]{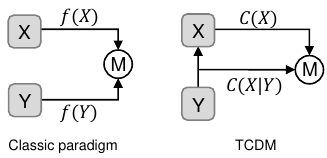}
    \caption{Paradigm of previous FR-PCQA metrics and TCDM \cite{zhang2023tcdm}.}
    \label{fig:tcdm}
    \vspace{-0.6cm}
\end{figure}

Considering that colored point clouds play a more important role in immersive media communication, researchers further propose a sequence of metrics that can measure color degradation. GraphSIM \cite{pcqa_yang2020_graphsim} is a pioneering work, which models point clouds as graph structures for effective feature extraction. 
As extensions of PC-MSDM and MSMD, PCQM \cite{pcqa_meynet2020_pcqm} and P2D \cite{javaheri2021point} additionally incorporate color-based features that consider lightness and contrast changes. PointSSIM \cite{alexiou2020towards} assesses a family of statistical dispersion measurements
for four types of attributes: geometry, normal vectors, curvature values, and
colors. However, the metric does not consider how to fuse the measurements of different types of attributes. Diniz \textit{et al.} propose a series of works \cite{diniz2020local,diniz2020towards,diniz2020multi,diniz2021color,diniz2022point} based on local binary patterns. In order to enhance the complementarity of the adopted features, TCDM \cite{zhang2023tcdm} (as shown in Fig.~\ref{fig:tcdm}) considers the FR-PCQA problem from another perspective, that is,
measuring the transformational complexity of recovering the reference from the distorted point clouds. The metric utilizes a space-aware vector autoregressive model to encode multiple
channels of the reference point cloud in cases with and without
its distorted version to obtain complexity- and prediction-based features for quality prediction.

To better adapt to human perception, researchers have taken into account the multi-scale characteristics when designing metrics. MS-GraphSIM \cite{zhang2021ms} constructs a multi-scale representation for local patches and then fuses GraphSIM
at different scales to obtain an overall quality score.  MS-PointSSIM \cite{lazzarotto2023towards} further produces a collection of PointSSIM scores for multiple scales obtained by voxelizing point clouds at different bit depth precisions. EPES \cite{xu2021epes} and \cite{xu2024compressed} both posit that global and local details contribute to the human perception and measure point cloud distortion based on global-local tradeoffs. MPED \cite{yang2022mped} regards point clouds as systems that have potential energy and the distortion can change the total potential
energy. By evaluating various neighborhood sizes, the metric can capture distortion in a
multi-scale fashion.

Some metrics have been proposed to utilize machine learning tools to address the FR-PCQA problem. VQA-CPC\cite{hua2020vqa}, FQM-GC\cite{zhang2021fqm}, and CPC-GSCT \cite{hua2022cpc} extract multiple geometry- and color-based features and then utilize random forest (RF) to perform feature pooling. GQI \cite{chetouani2021convolutional} defines a patch of size $32\times32$ around each
pair of points and leverages the convolutional neural network (CNN) to perform quality prediction. \cite{tliba2022point} measures
the cross-correlation between the embedding of reference and
distorted point clouds to quantify visual distortion. PointPCA \cite{alexiou2024pointpca,zhou2023pointpca+} and \cite{laazoufi20223d} both utilize a series of descriptors obtained from the eigendecomposition of the covariance matrix and then perform quality regression through RF. SGW-PCQA \cite{watanabe2024full, watanabe2025full} compares features between reference and distorted point clouds and then uses support vector regression (SVR) to predict the quality score.

To complement the drawbacks of structure-wise metrics, there is a trend to combine them with projection-wise or point-wise metrics. CF-PCQA \cite{cui2024colored} develops four features in the 3D space and extracts structural similarity
features and wavelet features from the projected 2D images. PHM \cite{zhang2024perception} adaptively leverages two visual strategies with respect to distortion degree to predict point cloud quality. The adaptive strategy results in a more robust prediction across different distortion environments.

\subsubsection{Projection-based Metrics}

The projection-based metrics rely on quantifying point cloud distortion using projected images. To simulate the multi-view characteristic when observing 3D objects, researchers usually project the point cloud onto 2D planes at different viewing angles. 

% \begin{itemize}

%     \item Paragraph1: Early projection-based metrics mostly utilize only texture images.

%     \item Paragraph2: To better gague the geometry deformation, some studies has incorporated geometry-related information for projection-based metrics. 

% \end{itemize}
Early projection-based metrics mainly focus on the use of texture images and investigate the impact of viewport, background color, and image quality assessment (IQA) metric selection. \cite{proj1, proj2, proj3} use six orthogonal projections of the bounding cube to project the point cloud onto the images and compare the average results of various IQA metrics across these six images. Alexiou \textit{et al.} 
\cite{proj_num_views} studies the influence of the number of views and the pooling strategy. The results indicate that increasing the number of views had limited benefits. Similarly, Liu \textit{et al.} \cite{proj_background} examine the impact of varying the number of views on the bounding sphere and find minimal differences. 
IW-SSIM is applied in \cite{proj_background} to exclude background information from the evaluation. Javaheri and Freitas \textit{et al.} \cite{Javaheri_Joint_Geometry,freitas2023point} propose innovative projection methods to address the issue of foreground mismatches.
To reduce the issue of holes in point cloud projections, SISIM \cite{sisim} proposes to scale the point cloud so that the average distance between points falls within a fixed range before projection.

\begin{figure}[t]
    \centering
    \includegraphics[width=0.9\linewidth]{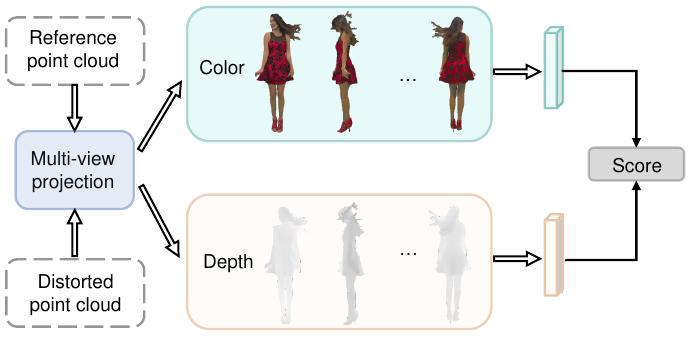}
    \caption{Paradigm of joint color and geometry metric\cite{proj_yang_depth}.}
    \label{fig:yangtmm}
    \vspace{-0.6cm}
\end{figure}

To better gauge geometric deformation and extract 3D-related features, some studies have incorporated geometry-related information and created handcrafted features into projection-based metrics. Yang \textit{et al.}\cite{proj_yang_depth} (as shown in Fig.~\ref{fig:yangtmm})  introduceadditional projected depth images to represent the geometric information of the point cloud. Chen \textit{et al.}
\cite{proj_layered} propose slicing the point cloud into layers and projecting each layer simultaneously.
SISIM \cite{sisim} leverages natural scene statistics to extract texture features and uses local binary patterns to capture geometric details. TGP-PCQA \cite{TGP_PCQA} employ 4D tensor decomposition to extract compressed texture information and uses features to characterize geometric properties.
Building on these ideas, Zhang \textit{et al.} \cite{zhang2023simple} employ neural networks to achieve feature extraction processes for improved performance.

\subsection{No-reference (NR) Point Cloud Quality Assessment}
\subsubsection{Model-based Metrics} 
NR model-based metrics directly map 3D data into qualtiy scores without available references.
 Based on the architectures employed for feature learning, they can be categorized into two primary groups: handcrafted-based and learning-based approaches. The formerfirst extracts handcrafted features (\textit{e.g.}, curvature, entropy, luminance) and then feeds them into CNN networks or SVR models to predict quality scores. The latter utilizes learnable CNN architectures to automatically extract features from raw point clouds or other data structures (\textit{e.g.}, graph).

For handcraft-based methods, Chetouani \textit{et al.} \cite{chetouani2021deep} extract three low-level handcrafted features (\textit{i.e.}, geometric distance, local curvature, and luminance values) from local patches. To enhance the extraction of handcrafted features, Zhang \textit{et al.} \cite{zhang2022no} first project the 3D models into geometry- and color-related feature domains for feature extraction and fuse features with a SVR model. Zhou \textit{et al.} \cite{zhou2024blind} propose a structure guided resampling method that resamples sparse keypoints using structural information and extracts three handcrafted features for quality regression. Additionally, MFE-Net \cite{liang2023mfe}  adaptively extracts handcrafted features from point cloud clusters. This method demonstrated improved flexibility and precision in  PCQA.

\begin{figure}[t]
    \centering
    \includegraphics[width=0.9\linewidth]{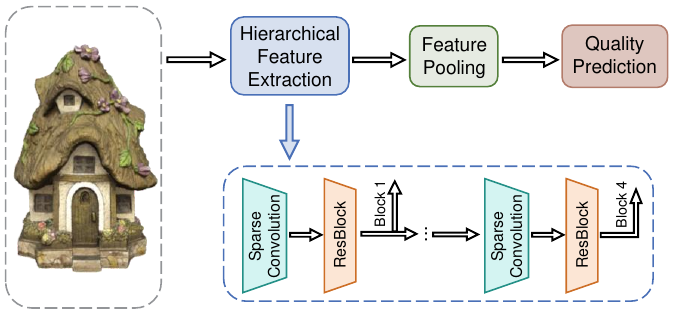}
    \caption{The framework of ResSCNN \cite{Liu2022ResSCNN}.}
    \label{fig:resscnn}
    \vspace{-0.6cm}
\end{figure}

For learning-based methods, some metrics are performed directly on raw 3D data.  Considering the huge volume of point cloud data and crucial structure information, ResSCNN\cite{Liu2022ResSCNN} converts point clouds into voxel representations and propose a sparse CNN (as shown in Fig.~\ref{fig:resscnn}) to extract the hierarchical features. \cite{tliba2022representation} learns a lightweight permutation invariant feature descriptor function, followed by a shallow quality regressor. Su \textit{et al.} \cite{su2022no} present the PRL-GQA framework, which integrates a Siamese network with a pairwise rank learning module. Xiong \textit{et al.} \cite{xiong2023psi} propose the PSI network, which consists of an elementary quality perception stream and a distortion perception stream. To expand the data size for training and testing, 3DTA \cite{zhu20243dta} designs a two-stage data sampling and a twin-attention-based transformer model.

Furthermore, to fully exploit the geometric characteristics of point clouds, other learning-based metrics first transform point clouds into 3D data structures (\textit{e.g.}, graph) for feature extraction. GPA-Net\cite{shan2023gpa} conduct pioneering work in this direction: they propose a multi-task graph convolutional network based on the improved graph kernel, which can attentively extract perturbations of structure and texture. Similarly, Tliba \textit{et al.}  \cite{tliba2023novel} propose a dual-stream architecture and utilize graph norm operation in each stream. WU \textit{et al.} \cite{wu2024no} present a method using structure sampling and clustering to capture global and local information using graph-based techniques. Inspired by the hierarchical perception system of human brains, PKT-PCQA\cite{liu2022progressive} converts the coarse-grained quality classification knowledge to the fine-grained quality prediction task. Considering the relationships between non-local points,  Wang \textit{et al.} \cite{wang2023non} propose a non-local geometry and color gradient aggregation graph model to evaluate the perceptual quality.

\subsubsection{Projection-based Metrics}
Similar to projection-based FR metrics, many NR metrics also project distorted point clouds into multi-view images to leverage 2D networks for quality score prediction.

Early projection-based metrics tend to project point cloud into textured images or videos to conveniently leverage CNN-like networks or 2D Transformers. \cite{liu2021pqa,bourbia2022blind,bourbia2023no} use CNNs to encode the projected image of the distorted point cloud under each view and predict the quality score by fusing the encoded features via symmetric pooling strategies such as averaging and maxpooling. 
Specifically, PQA-Net \cite{liu2021pqa} additionally trains a head for distortion type classification to equip the network with useful distortion-related prior knowledge.  
To mimic the dynamic viewpoints of humans when carefully observing point clouds, \cite{fan2022no, zhang2023evaluating} evaluate the point clouds from moving camera videos and predict scores using video quality assessment (VQA) methods. These metrics use temporal feature extraction to implicitly model the correlation between projected images of different views. 
Other than temporal modeling, MVAT \cite{mu2023multi} and  GC-PCQA \cite{chen2024no} use an aggregation transformer or a spatial graph to characterize the mutual dependencies and interactions between different viewpoints.  Fan \textit{et al.} 
\cite{fan2024uncertainty} propose a novel probabilistic architecture to model the judging stochasticity of subjects.
Furthermore,  Zhang \textit{et al.} \cite{zhang2024optimizing} explore uneven viewpoint selection for projection-based metrics according to point clouds' visible-points ratio and visible-color-entropy ratio. 
To reduce the high complexity when simultaneously processing multi-view images, Zhang \textit{et al.} \cite{zhang2023eep,zhang2024gms} utilize the mini patch sampling strategy to spatially downsample the projected images from the six perpendicular viewpoints.

% \begin{figure}[t]
%     \centering
%     \includegraphics[width=0.5\textwidth]{image/pipeline_projection_nr.png}
%     \caption{Framework of Projection-based PQA-Net.}
%     \label{fig:nr-projection}
% \end{figure}
To effectively capture quality information, many projection-based metrics resort to multi-scale feature extraction because point cloud quality is determined by high-level content and low-level distortion. MOD-PCQA \cite{wang2024zoom} uses a multi-scale feature extraction framework to capture both fine and coarse visual information for enhanced quality prediction. 
MS-PCQE \cite{chai2024ms} integrates varying focal lengths into a convolutional gated recurrent unit module. Miao \textit{et al.}  
\cite{miao2024no} use a hierarchical pyramid network to dissect point clouds into distinct layers, enabling the extraction of features across various scales.

To accurately capture geometric distortions, many studies incorporate geometric information, such as depth map, into projection-based metrics. PM-BVQA \cite{tao2021point} projects colored point cloud into texture and geometric projection maps, and fuses their latent features to capture complementary information.
Similarly, Tu \textit{et al.} \cite{tu2022v} projects point cloud 2D texture and geometry maps using V-PCC, extracts and fuses global and local features via a dual-stream network.
\cite{bourbia2022no, zhou2024visual,zhou2023bpqa} leverage saliency maps to weight each projected image. Xie \textit{et al.}  \cite{xie2023pmbqa} comprehensively project point clouds into normal, depth, and roughness maps and develop a deformable convolution-based alignment module.
Furthermore, AFQ-Net \cite{zhang2024asynchronous} uses global attention maps to guide feature fusion of texture and depth maps in a coarse-to-fine manner. The dynamic convolution operation is employed on different semantic regions to obtain the local feature, followed by a coarse-to-fine quality prediction.

The aforementioned projection-based metrics rely on labels collected from the PCQA databases for training. However, labels are limited in scale because of the time-consuming and labor-intensive collection process, resulting in poor generalizability when encountering unseen content or distortions. To address this challenge, many projection-based metrics leverage additional training data to address the label scarcity problem and boost generalizability. IT-PCQA \cite{yang2022no} leverages large-scale IQA datasets through unsupervised adversarial domain adaptation, which transfers knowledge from images to point clouds by projecting point clouds into multi-view images. 
\cite{tliba2024balancing} leverages self-supervised learning to address label scarcity in point cloud quality assessment by introducing quality-driven self-supervised loss. 
CoPA \cite{shan2024contrastive} improves the classical contrastive learning paradigm to generalize to quality assessment tasks. CoPA utilizes local patch mixing to retain distortion patterns when generating anchors, and divides negative samples into two parts to distinctly learn distortion-aware and content-aware representations.
PAME \cite{shan2024pame} and DisPA \cite{shan2025learning} both utilize masked autoencoding (MAE) to learn quality-aware representations with unlabeled point cloud data. Furthermore, DisPA uses a mutual information minimization strategy to enhance the discrepancy between  content and distortion representations.
$\text{D}^3\text{-PCQA}$ \cite{liu2023once} addresses the problem of label scarcity by using domain-relevance degradation description to establish an intermediate description domain.

\subsubsection{Multi Modal-based Metrics}
 % \textbf{Metrics that utilize 2D and 3D information. MM-PCQA} 
Due to the variety of distortion types, relying solely on a single modality is inherently limited and struggles to effectively address complex distortion scenarios. Consequently, some studies have advanced the field by incorporating multi-modal information.

According to the observation that the 3D point
cloud modality is more sensitive to geometry deformations, while the 2D projected image modality is sensitive to color distortions, some works propose to fuse 3D and 2D information.
Early metrics utilize handcrafted features followed by traditional prediction model ({\textit{e.g.}}, RF and SVR) to assess the quality.
BQE-CVP \cite{hua2021bqe} simultaneously extracts geometric features from segmented regions of the point cloud and color features from both point clouds and projected images. 
Tu \textit{et al.} \cite{tu2023pseudo} construct a pseudo-reference point cloud for the 3D modality and create both a texture projection map and a geometry projection map for the 2D modality.
Li \textit{et al.} \cite{li2025robust} assumes that an expected 3D scan will have a uniformly distributed point cloud on surfaces, and analyze the quality of points by using geometric information from surfaces fitted to these points.

\begin{figure}
    \centering
    \includegraphics[width=0.9\linewidth]{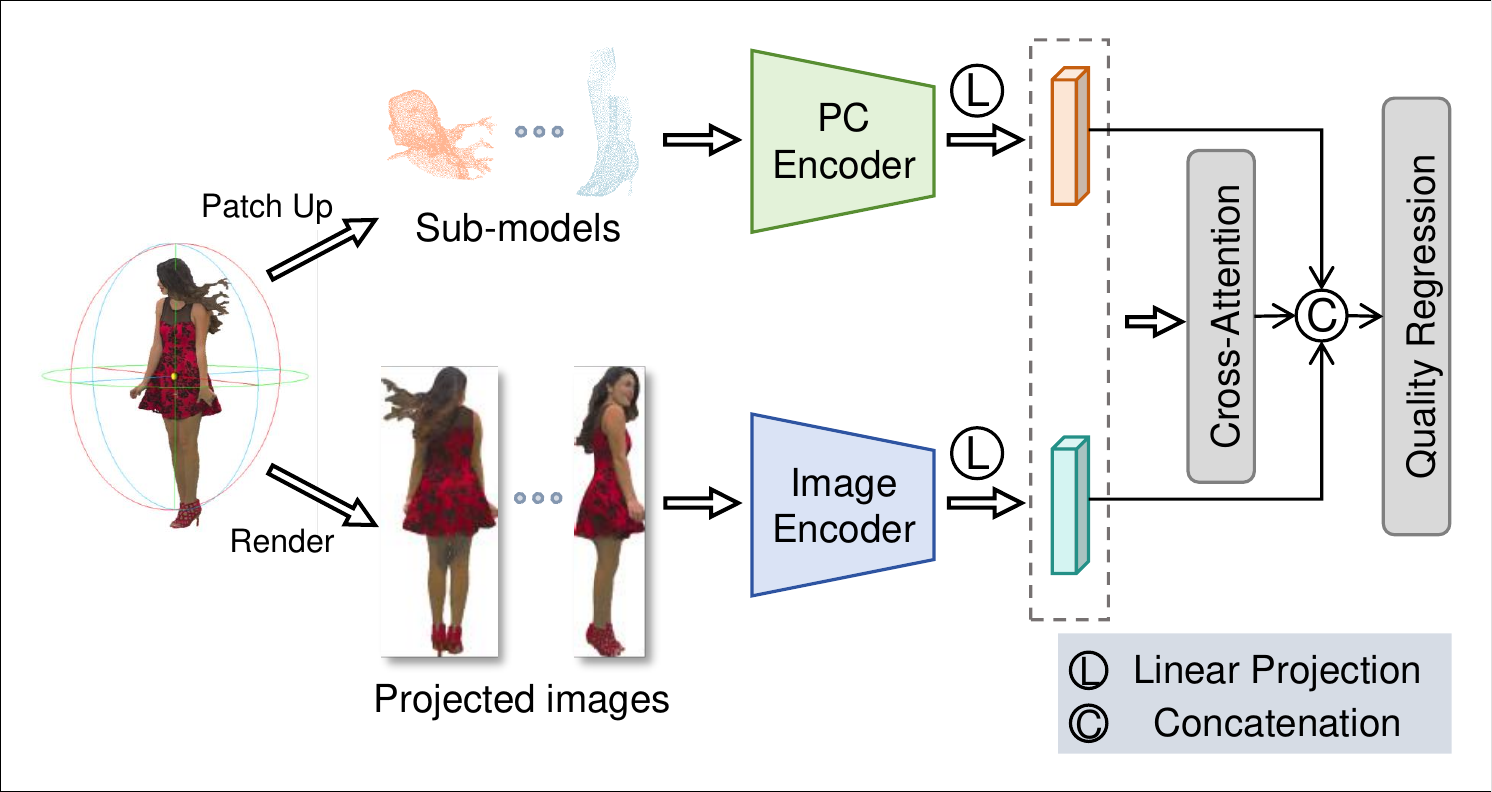}
    \caption{The framework of MM-PCQA~\cite{zhang2022mm}.}
    \label{fig:mmpcqa}
    \vspace{-0.6cm}
\end{figure}

\begin{table*}[t]
\centering
\caption{Performance of FR-PCQA metrics.  The top two performance results are marked in \textbf{boldface} and \underline{underline}.}
\label{tab:overall_performance_FR}
\resizebox{0.9\textwidth}{!}{
\begin{tabular}{c|c|ccc|ccc|ccc|ccc}
\toprule
\multirow{2}{*}{Metric} &\multirow{2}{*}{Type} & \multicolumn{3}{c|}{SJTU-PCQA\cite{proj_yang_depth}}  & \multicolumn{3}{c|}{WPC\cite{proj3}} & \multicolumn{3}{c|}{LS-PCQA\cite{Liu2022ResSCNN}} & \multicolumn{3}{c}{BASICS\cite{ak2024basics}}  \\ \cline{3-14}
& & PLCC & SROCC & RMSE & PLCC & SROCC & RMSE & PLCC & SROCC & RMSE & PLCC & SROCC & RMSE  \\ \midrule

% Joint &I
% \\ \midrule

MSE-p2po-PSNR  \cite{pcqa_p2p} &M(P) &0.877	&0.791	&1.166	&0.578	&0.566	&18.708	&0.487	&0.302	&0.727	 &0.633	&0.485	&0.829

\\
Haussdorf-p2po-PSNR \cite{pcqa_p2p} &M(P) &0.742	&0.681	&1.628	&0.398	&0.258	&21.028	&0.403	&0.269	&0.761	&0.190	&0.250	&1.051

\\
MSE-p2pl-PSNR \cite{pcqa_p2p} &M(P) &0.753	&0.676	&1.596	&0.488	&0.446	&20.013	&0.444	&0.287	&0.745	&0.627	&0.516	&0.834

\\
Haussdorf-p2pl-PSNR\cite{pcqa_p2p} &M(P) &0.737	&0.670	&1.639	&0.383	&0.315	&21.171	&0.401	&0.269	&0.762	&0.230	&0.284	&1.042

\\
$\rm PSNR_{YUV}$ \cite{pcqa_p2p} &M(P) &0.652	&0.646	&1.841	&0.551	&0.536	&19.132	&0.497	&0.476	&0.722	&0.603	&0.585	&0.853

\\ 
\midrule
pl2pl  \cite{alexiou2018point} &M(S) &0.670	&0.482	&1.801	&0.325	&0.303	&21.675	&0.231	&0.199	&0.809	&0.121	&0.100	&1.063
 \\
 MSMD \cite{javaheri2020mahalanobis} &M(S) &0.630 &0.612	&1.885 &0.487 &0.430 &20.022 &0.566 &0.548 &0.686 &0.794 &0.683 &0.651
\\
% P2I  &M(S) \\
PCQM \cite{pcqa_meynet2020_pcqm} &M(S) &0.860	&0.847	&1.237	&0.751	&0.743	&15.132	&0.346	&0.321	&0.780	&\underline{0.888}	&0.810	&\underline{0.492}

\\
GraphSIM \cite{pcqa_yang2020_graphsim} &M(S) &0.856	&0.841	&1.071	&0.694	&0.680	&16.498	&0.355	&0.332	&0.778	&\bf{0.893}	&\bf{0.818}	&\bf{0.481}

\\
MS-GraphSIM \cite{zhang2021ms} &M(S) &0.897	&0.874	&1.071	&0.717	&0.707	&15.97	&0.432	&0.404	&0.750	&-	&- &- 
\\
pointSSIM \cite{alexiou2020towards} &M(S)&0.725	&0.704	&1.672	&0.510	&0.454	&19.713	&0.291	&0.157	&0.796	&0.725	&0.692	&0.737

\\
MPED \cite{yang2022mped} &M(S) &0.896	&0.884	&1.076	&0.700	&0.678	&16.374	&\bf{0.648}	&\bf{0.601}&\bf{0.634}	&0.803	&0.708	&0.638

\\ 

TCDM \cite{zhang2023tcdm} &M(S) &\bf{0.930}	&\bf{0.910}	&\bf{0.891}	&\underline{0.807}	&\underline{0.804}	&\underline{13.525}	&0.433	&0.408	&0.750	&0.859	&0.740	&0.547

\\
PHM \cite{zhang2024perception} &M(P+S) &\underline{0.907}	&\underline{0.888} 	&\underline{1.021}	&\bf{0.839}	&\bf{0.833}	&\bf{12.466}	&\underline{0.615}	&\underline{0.589}	&\underline{0.656} &0.872	&\underline{0.817}	&0.525

\\ \bottomrule
\end{tabular}}
\end{table*}

\begin{table*}[t]
\centering
\caption{Performance of NR-PCQA metrics.  The top two performance results are marked in \textbf{boldface} and \underline{underline}.}
\label{tab:overall_performance_NR}
\resizebox{0.9\textwidth}{!}{
\begin{tabular}{c|c|ccc|ccc|ccc|ccc}
\toprule
\multirow{2}{*}{Metric} &\multirow{2}{*}{Type} & \multicolumn{3}{c|}{SJTU-PCQA\cite{proj_yang_depth}}  & \multicolumn{3}{c|}{WPC\cite{proj3}} & \multicolumn{3}{c|}{LS-PCQA\cite{Liu2022ResSCNN}} & \multicolumn{3}{c}{BASICS\cite{ak2024basics}}  \\ \cline{3-14}
& & PLCC & SROCC & RMSE & PLCC & SROCC & RMSE & PLCC & SROCC & RMSE & PLCC & SROCC & RMSE  \\ \midrule

ResSCNN \cite{Liu2022ResSCNN} &M &0.889	&0.880	&0.878	&0.817	&0.799	&13.082	&0.648	&0.620	&0.615 &0.391	&0.352	&0.975

\\
GPA-Net \cite{shan2023gpa} &M  &0.901	&0.891	&0.864	&0.796	&0.775	&12.976	&0.623	&0.602	&0.705 &0.391 &0.352 &0.975
\\ 
3DTA \cite{zhu20243dta} &M  &0.948	&\textbf{0.931}	&
0.910	&\underline{0.886}	&\underline{0.886}	&\underline{11.636}	&0.609	&0.604	&0.660 &0.896	&0.825	&0.488\\ 

\midrule
PQA-Net \cite{liu2021pqa} &I &0.898	&0.875	&1.340	&0.789	&0.774	&12.224	&0.607	&0.595	&0.697 &0.655	&0.387	&0.797

\\

IT-PCQA \cite{yang2022no} &I &0.609	&0.602	&1.559	&0.581	&0.562	&13.889	&0.447	&0.423	&0.899 &0.367 &0.344 &0.997
\\

VQA-PC \cite{zhang2023evaluating}  &I & 0.870 & 0.861 & 1.101 & 0.800 & 0.801 & 13.558 & - & - & - & - & - & -  
\\
MOD-PCQA \cite{wang2024zoom} &I & 0.953 & 0.931 & 0.712 & 0.873 & 0.875 & 11.060 & - & - & - & - & - & - 
\\
MS-PCQE \cite{chai2024ms} &I & 0.933 & 0.918 & 0.824 & 0.875 & 0.874 & 11.028 & - & - & - & - & - & - 
\\
CoPA\cite{shan2024contrastive} &I & 0.912 & 0.885 & 0.988 & 0.787 & 0.754 & 13.588 & 0.625 & 0.609 & 0.646 & 0.754 & 0.730 & 0.512
\\

GMS-3DQA \cite{zhang2024gms} &I  &0.916	&0.886	&0.931 &0.818	&0.814	&13.032 &0.666 &0.645	&0.606 &0.895	&\underline{0.807}	&0.472

\\

AFQ-Net \cite{zhang2024asynchronous} &I   &\bf{0.957}	&{0.930}	&\bf{0.678}	&\bf{0.897}	&\bf{0.899}	&\bf{9.971}	&\underline{0.690}	&\underline{0.680} &\underline{0.583} &\underline{0.905} &0.793 &\underline{0.449}
\\
\midrule
MM-PCQA \cite{zhang2022mm}  & M+I 
 &0.939	&0.910	&0.805	&0.846	&0.844	&12.028	&0.644	&0.605	&0.621 &0.793	&0.738	&0.628

\\ 
% MM-PCQA  & M+I 
%  &0.939	&0.910	&0.805	&0.846	&0.844	&12.028	&0.644	&0.605	&0.621 &0.793	&0.738	&0.628

% \\ 
DHCN \cite{chen2024dynamic} &M+I & \underline{0.957} &\bf{0.942} &\underline{0.681} &0.866 &0.862&11.391 & - & - & - & - & - & -
\\
M3-Unity \cite{zhou2024deciphering} &M+I &0.925 &0.885 &0.887 &0.839 &0.837 &12.304 &0.646 &0.631 &0.625 &0.843 &0.790 &0.562\\
LMM-PCQA \cite{zhang2024lmm} &I+L & 0.940 & \underline{0.938} &0.718 & 0.874 & 0.883 & 11.817 & - & - & - & - & - & -\\
CLIP-PCQA \cite{liu2025clip} &I+L &0.956 &0.936 &0.693 &\underline{0.894} &\underline{0.890} &\underline{10.112} &\bf{0.755} &\bf{0.736} &\bf{0.533} &\bf{0.932} &\bf{0.872} &\bf{0.382}

\\ \bottomrule
\end{tabular}}

\vspace{-0.6cm}
\end{table*}

Recent learning-based metrics strengthen 3D and 2D feature representation based on end-to-end learning paradigm. 
As a pioneering work, MM-PCQA\cite{zhang2022mm} (as shown in Fig.~\ref{fig:mmpcqa}) uses independent 2D and 3D backbones to extract latent features and simply fuses them before quality regression. 
M3-Unity \cite{zhou2024deciphering}  designs a multi-task decoder that selects the best combination among four modalities based on the distortion type.
RAL-net \cite{wang2024rating} is also a multi-task framework concerning both quality score regression and quality level classification. 
    
Some subsequent work places emphasis on more effectively extracting and leveraging features from these two modalities.
MHyNet-PC\cite{chatterjee2024mhynet}  reduces computational complexity by extracting 3D geometric features through statistical analysis, while leveraging deep learning to extract features from 2D projections.
Hallucinated-PQA \cite{mu2024hallucinated} fuses multi-view features and high-level geometric semantic features.
Another line of research primarily aims to investigate the interaction and integration of multimodal data. Chen \textit{et al.} \cite{chen2024dynamic} design a dynamic hypergraph convolutional network for feature fusion, using a dynamic hypergraph generator to capture the correlation between two modalities.
MFT-PCQA \cite{liu2024mft} propose a mediate-fusion strategy to exploit self-learning of each modality and cross-learning within modalities simultaneously.
Ding \textit{et al.} \cite{ding2024point} design a cross-modal de-redundancy module, which first reduces the information redundancy between features and then performs attention computation to explore the information between different modalities.
MMF-PCQM\cite{wang2023applying} introduces a cross-modal collaborative adversarial learning strategy and a fine-grained feature fusion module under a dual-branch structure. 
Other methods both innovate feature extraction and explore more effective multi-modal fusion.
Wave-PCT \cite{guo2024wave}  applies wavelet transforms to point cloud patches and projected images to extract multi-scale local spectral features. 
Plain-PCQA \cite{chai2024plain} uses a three-branch network architecture to extract anchor-based visual features, exploring the plane-point (2D-3D) interactive mechanism, respectively.
CMDC-PCQA \cite{wu2025cmdc} uses both point space and graph structure to extract point cloud features, and utilizes a cross-modal deep-coupling framework to emphasize the global and local weights between features of different modalities.
    
    % \item \textbf{Metrics that incoporate language knowledge.  CLIP-PCQA, LMM-PCQA}
    
With the development of vision-language learning and large language models (LLM), some novel metrics leverage the capacity of natural language to understand visual concepts for PCQA task.
CLIP-PCQA \cite{liu2025clip} leverages quality descriptions to characterize  intuitive human perception and align with subjective experiments by predicting the opinion score distribution.
LMM-PCQA \cite{zhang2024lmm} instructs LMM with question-answer pairs regarding PCQA issues, and quantifies geometry distortions using the estimation of key statistic parameters to help LMM gain a more comprehensive understanding of point cloud quality.
\cite{xie2024llm} uses LMM to obtain relevant quality descriptions. 
\cite{guptapit} present an end-to-end point-image-text multi-modal model PIT-QMM, which bridges the algorithmic gap between 2D and 3D quality assessment.

% \subsubsection{Bitstream-based Metrics}

% Bitstream-based metrics has aimed to measure point cloud quality using payload information extracted from the compressed bitstream, which benefits real-time and nonintrusive quality monitoring. 

% Dumic et al. \cite{Bitstreamsubjective} study the effect of packet losses on the subjective quality of V-PCC-compressed dynamic point clouds. 
% Cao et al. \cite {BitstreamMeshPointCloud} compare point cloud and mesh sequence compression to determine the preferred representation based on content, bit rate, and observation distance. They also propose a subjective rating model with two components: a bit rate quality factor (BQF) for compression quality based on bit rate, and an observation distance correction factor to adjust BQF for distance.

% Several authors have designed bitstream-based quality assessment methods tailored to specific encoding methods. 
% StreamPCQ-OL, StreamPCQ-OR and StreamPCQ-TL \cite{BitstreamOctree-Lifting, BitstreamOctree-RAHT, BitstreamTrisoup-Lifting} is designed for different compression mode of MPEG codec, linking subjective quality degradation to texture quantization parameters (TQP) and texture complexity (TC). 
% Liu et al. \cite{BitstreamVPCC} propose a model for V-PCC encoded point clouds, where perceptual distortion depends on TC estimated using TQP and texture bitrate per pixel (TBPP).
% streamPCQ-OR 

\subsection{Experiments}

\subsubsection{Databases and Evaluation Protocols}
\ 
\newline
\indent \textbf{Database.} 
To comprehensively evaluate point cloud quality assessment metrics, we employ four widely recognized benchmarks: SJTU-PCQA \cite{proj_yang_depth}, WPC \cite{proj3}, LS-PCQA \cite{Liu2022ResSCNN}, and BASICS \cite{ak2024basics}. These datasets represent diverse point cloud distortion scenarios and subjective evaluation methodologies. 

The SJTU-PCQA dataset comprises a collection of nine high-fidelity reference point clouds accompanied by 378 degraded variants under four fundamental impairments and three compound distortions.
WPC contains 20 reference point clouds and 740 distorted point clouds generated from the references under five types of distortions, including downsampling (DS), Gaussian noise contamination (GNC), G-PCC(Trisoup), G-PCC(Octree), and V-PCC. 
The LS-PCQA dataset contains 85 reference point clouds and 930 corresponding distorted samples created using 31 different degradation methods. 
The BASICS dataset contains 75 high-quality reference point clouds along with 1494 distorted variants produced through the application of four distinct compression techniques at various configuration levels.

\textbf{Evaluation Protocols.} 
The effectiveness of objective quality assessment models can be quantified by their alignment with human perceptual judgments across three critical dimensions: accuracy, monotonicity, and consistency. To evaluate these aspects, researchers typically employ three standard performance indicators: Spearman rank-order correlation coefficient (SROCC), Pearson linear correlation coefficient (PLCC), and root mean squared error (RMSE). The superior performance of the model is indicated by the SROCC and PLCC values approaching 1.0, coupled with the RMSE values approaching zero. It should be noted that before calculating PLCC and RMSE, a four-parameter logistic regression is typically applied to transform objective predictions onto the same scale as Mean Opinion Scores (MOS).

\textbf{Experiment Setup.} 
Representative models from each category of FR and NR point cloud quality assessment algorithms, as previously introduced, are selected for performance
comparison on datasets having diverse distortion types. For FR-PCQA metrics, we directly report their performance on the whole databases; for NR-PCQA metrics, the k-fold cross-validation strategy used in \cite{zhang2024asynchronous}  is employed for the experiment. Specifically, 9-fold is selected for SJTU-PCQA (9 references). For WPC (20 references), LS-PCQA (85 references), and BASICS (75 references), we apply a 5-fold cross-validation.  Where possible, we retrain and evaluate the models by ourselves; for remaining cases, performance statistics are extracted directly from the original publications. 
The performance comparison results of FR models and NR models are summarized in Table~\ref{tab:overall_performance_FR} and ~\ref{tab:overall_performance_NR}, respectively.
\subsubsection{Performance Comparison for FR-PCQA}
As shown in Table~\ref{tab:overall_performance_FR}, early point-based and projection-based metrics provide good performance on some databases, such as MSE-p2po-PSNR on SJTU-PCQA. However, due to the lack of structural information or geometry characteristics, they generally provide inferior results on those databases with more content and distortion types such as LS-PCQA and BASICS.

By incorporating information from a macro scale, structure-based metrics generally provide better performance on these databases. MS-GraphSIM is superior to GraphSIM in most cases, demonstrating the importance of multi-sale property. MPED works well on LS-PCQA because it extracts features from relatively small regions (\textit{e.g.}, five nearest neighbors), thus maintaining the sensitivity to microscopic noises that are prevalent on LS-PCQA. Notably, although some metrics provide outstanding results in certain instances, they perform less well in other cases (\textit{e.g.}, TCDM on LS-PCQA and MPED on M-PCCD). PHM alleviates this problem by leveraging hybrid strategies with respect to distortion degree to predict point cloud quality, showing good generalization for different distortion environments.

We offer the following design insights for the FR-PCQA metrics:

\begin{itemize}
    \item Multi-scale modeling. Multi-scale perception plays an important role in human vision. We observe significant performance improvements of multi-scale metrics over single-scale metrics, such as MS-GraphSIM \textit{v.s.} GraphSIM. The weighting of different scales is also an important point to consider.
    \item Complementary feature extraction. The success of quality metrics is highly dependent on the effectiveness of the extracted features. Due to the irregular 3D structure of the point cloud and rich distortion types, a single feature is usually insufficient to handle various types of distortion. Consequently, some metrics like TCDM   
    efficiently and comprehensively model point cloud quality by extracting complementary features.
    \item Hybrid strategy. Whether it is point-based, structure-based or projection-based metrics, they can only characterize part of perceptual properties. Injecting hybrid strategies like PHM can effectively enhance the generalization towards different distortion environments.

\end{itemize}

\subsubsection{Performance Comparison for NR-PCQA}
As illustrated in Tab.~\ref{tab:overall_performance_NR}, deep learning-based NR metrics, by incorporating deep features and end-to-end training, demonstrate exceptional performance across various databases.

Among model-based metrics, ResSCNN and GPA-Net rely on local geometric features, which limits their generalization capabilities. Although they perform well on the SJTU-PCQA, WPC, and LS-PCQA databases, their performance on BASICS is notably poor. Conversely, 3DTA's two-stage sampling strategy and transformer-based architecture enhance data utilization and long-range dependencies, resulting in superior performance.

Among image-based metrics, early approaches like IT-PCQA, which employs less sophisticated domain adaptation techniques, underperform across all databases. Subsequent metrics achieve significant performance improvements by exploring the integration of local and global information, multi-scale feature extraction, and inter-perspective correlations. In particular, AFQ-Net emerges as the top performer, delivering strong results across all databases.
Despite the inherent information loss of image-based methods, their performance is competitive with model-based methods. This is primarily due to their proficiency in extracting texture information, whereas model-based metrics tend to focus more on geometric features and may not fully exploit comprehensive features.

Multi-modal metrics exhibit high performance by leveraging complementary information across multiple modalities. Early metrics such as MM-PCQA and M3-Unity employ simple feature fusion strategies, which prove to be less effective. However, DHCN improves feature extraction and interaction mechanisms, leading to improved performance. Notably, two metrics that introduce language modalities (LMM-PCQA and CLIP-PCQA) establish interpretable quality semantic spaces by aligning quality descriptive texts with subjective scores,  boosting model performance and generalization ability. 

We provide some design insights for deep learning-based NR-PCQA metrics as follows:

\begin{itemize}
    \item Advanced feature extraction strategies. The effectiveness of NR metrics heavily relies on the quality of the extracted features. Future metrics could explore innovative feature extraction techniques, such as hierarchical learning or attention-based mechanisms, to better capture the complex characteristics of point clouds. 
    \item Integration of local and global information. Local features capture fine-grained details, while global features provide a holistic view. Metrics that balance both aspects, such as AFQ-Net, have shown superior performance improvement. Future metrics could focus on designing mechanisms to effectively integrate different information.
    \item Multi-modal information fusion. The success of metrics like DHCN and CLIP-PCQA demonstrates the potential to integrate diverse modalities. Future metrics could explore more effective fusion techniques to combine information from different modalities enhance the understanding of point cloud distortions. 

\end{itemize}

\section{Applications and challenges}\label{sec:application}
\subsection{Compression of Digital Humans Based on Point Clouds}
High-fidelity digital human point clouds have emerged as a widely used representation due to their simplicity, flexibility, and suitability for real-time user engagement applications. Methods have been explored to enable efficient storage and transmission of digital humans: Subramanyam \textit{et al.} \cite{pcc_app1} evaluate the visual quality of V-PCC and G-PCC, and indicate that V-PCC generally outperforms G-PCC at lower bitrates but neither achieves lossless reconstruction at higher bitrates. To enhance semantic preservation, Xie \textit{et al.} \cite{pcc_app2} propose a region-of-interest (ROI) guided geometry compression method, while Wu \textit{et al.} \cite{pcc_app3} incorporate human geometric priors to improve compression efficiency.

Despite these advancements, several technical challenges remain in the compression of digital human point clouds: 1) Poor reconstruction quality. Compression artifacts and geometry degradation are particularly pronounced in regions requiring high detail, such as facial features and clothing, leading to visibly unrealistic reconstructions. 2) Limited scalability. Existing compression methods often struggle with high-resolution or temporally dynamic point clouds, making them less suitable for real-time applications such as VR/AR, which demand both high visual fidelity and low latency.

\subsection{LiDAR Point Cloud Compression for Autonomous Driving}
The advancement of autonomous driving relies on high-throughput sensing systems for real-time environmental perception. However, the large volume of LiDAR data, often hundreds of megabytes per second, presents challenges for onboard storage, low-latency processing, and wireless transmission. To manage this data deluge, efficient LiDAR point cloud compression has become a pressing requirement. Feng \textit{et al.} \cite{pcc_lidar1} leverage spatial and temporal redundancies across sequential point cloud frames, achieving compression rates ranging from 40× to 90×. Zhao \textit{et al.} \cite{pcc_lidar3} propose a deep learning-based approach utilizing bi-directional frame prediction with an asymmetric residual module and 3D space-time convolutions to achieve higher compression efficiency. 

Although these methods improve compression performance, the primary challenge is real-time performance. Advanced techniques, especially deep learning-based ones, often incur significant computational overhead, causing delays in encoding and decoding that compromise the responsiveness and safety of autonomous systems.
\subsection{Perceptual Optimization for Point Cloud Enhancement}

Visual enhancement technologies including denoising, upsampling, completion can significantly enhance perceptual
quality of point clouds. PCQA metrics can serve as loss functions to guide model optimization in learning-based point cloud enhancement tasks. There are two mainly used quality quantifications in early point cloud enhancement tasks: Chamfer Distance (CD) and Earth Mover's Distance (EMD). The former is low resource
but usually encounters problematic point matching; in contrast, the latter is stricter when performing point matching but has expensive computational cost. To better guide the optimization, Wu \textit{et al.} \cite{ wu2021density} propose density-aware CD to better detect the disparity of density distributions. Liu \textit{et al.} \cite{liu2020morphing} provide a lightweight version of EMD with lower complexity. \cite{urbach2020dpdist}  proposes DPDist that compares point clouds by measuring the distance between the surfaces that they were sampled on. Yang \textit{et al.} \cite{yang2022mped} present MPED which has similar computational complexity to CD and can better capture structural distortions.

Despite their utility, PCQA metrics face several challenges: 1) Subjective gaps. Existing metrics may not align with
perceptual quality when measuring restoration results, requiring large-scale subjective studies to benchmark
and calibrate metrics. 2) Lack of color optimization. Existing metrics mostly focus on evaluating geometry fidelity while overlooking color information. 3) Computational costs. Deep learning-based metrics
are resource-intensive, limiting real-time applications. Developing
adaptable metrics remains a significant challenge.

\vspace{-0.4cm}

\subsection{AI-Generated Content 3D Quality Assessment}

Rapid advances in machine learning have driven significant progress in AI-generated content (AIGC) \cite{liu2024comprehensive}. Previous works have conducted 3D AIGC evaluation by proposing new benchmarks, and applying reinforcement learning from human feedback (RLHF) to generative models. Benchmarks such as GPTEval3D \cite{wu2024gpt4v}, MATE-3D \cite{zhang2024benchmarking}, $\mathrm{T}^3$Bench \cite{he2023t3bench}, and 3DGCQA \cite{zhou20243dgcqa} have emerged to assess key aspects like text-to-3D alignment, geometry, texture, detail, and overall quality. Some researchers recognize the effectiveness of RLHF in improving the performance of generative models. DreamReward \cite{ye2024dreamreward}, DreamDPO \cite{zhou2025dreamdpo}, and other metrics \cite{liu2025improving} greatly boost high-text alignment and high-quality text-to-3D generation through human preference feedback.

However, these advancements also introduce distinct challenges. 1) Incomprehensive benchmarks. Although existing benchmarks \cite{wu2024gpt4v, zhang2024benchmarking, he2023t3bench, zhou20243dgcqa} have classified prompts based on textual complexity or object quantity, their scale is relatively small, and the evaluation perspectives are not comprehensive. 2) Ineffective measures. Unlike conventional distortions in 3D data processing, AIGC content exhibits unique degradation issues, including text-semantic misalignment and multi-view inconsistency (\textit{i.e.}, the Janus problem) \cite{hong2023debiasing,dreamfusion}, which pose new challenges for developing objective metrics. 3) Insufficient guidance. 3D content generated by existing metrics often does not fully align with human preferences. Despite the proven efficacy of reward models \cite{ye2024dreamreward,zhou2025dreamdpo}, learning from human feedback still requires investigation.

\section{Conclusion}\label{sec:conclusion}
The massive volume and unique characteristics of point clouds emphasize the importance of efficient compression and quality assessment algorithms. This paper provides a comprehensive overview of advancements in point cloud compression and quality assessment across multiple datasets, highlighting their critical roles in various applications such as immersive applications and robotics. Although significant progress has been made, several challenges remain, including improving visual fidelity, reducing latency, exploring color optimization, and expanding to emerging content. Future research should incorporate techniques like hybrid frameworks, along with advanced and multi-modal feature extraction, enabling more immersive, efficient, and intelligent 3D applications.

% % \ifCLASSOPTIONcaptionsoff
% %   \newpage
% % \fi

\bibliographystyle{IEEEtran}
\bibliography{ref}

\end{document}